\pdfoutput=1

\documentclass[11pt]{article}

\usepackage[final]{acl}

\usepackage{times}
\usepackage{latexsym}

\usepackage[T1]{fontenc}
\usepackage[utf8]{inputenc}

\usepackage{microtype}

\usepackage{inconsolata}

\usepackage{booktabs}
\usepackage{longtable}
\usepackage{multirow}
\usepackage{siunitx}
\usepackage{svg}
\usepackage{xcolor}
\usepackage{subcaption}

\title{Auditing Exposure to Harmful Content on TikTok using Multimodal Language Models:\\
A Cross-National, Age-Stratified Study}

\author{Hamidreza Saffari\textsuperscript{1} \and Francesco Pierri\textsuperscript{1} \\
  \textsuperscript{1}Politecnico di Milano \\
  \texttt{hamidreza.saffari@mail.polimi.it}, \texttt{francesco.pierri@polimi.it}
}

\begin{document}
\maketitle

\begin{abstract}
Online video platforms can expose young users to harmful content, but independent audits remain difficult because video annotation is costly and moderation judgments vary across languages.
We audit TikTok in France, Italy, and Sweden with sockpuppet accounts representing four age personas (13, 16, 19, 40), collecting $36{,}971$ videos from passive For-You-page scrolling and active sessions that scroll, search for harm keywords, and scroll again.
To scale annotation, we validate four multimodal LLMs against native-speaker labels on a 300-video reference set.
Gemini 2.5 Flash with eight sampled frames plus text performs best (aggregate $\kappa = 0.42$), at half the per-call cost of native-video upload, and we apply it to a $10\%$ sample for approximately \$50 in total API spend across both modalities.
Keyword search returns $35$--$56\%$ harmful content, a $1.5$--$7.5\times$ increase over the scrolling baseline in ten of twelve country--age combinations; the spike is temporary and flattens the age differences observed in France and Sweden.
Under passive scrolling, Italy has the highest harm rate at every age, with Italian age-19 reaching $48.6\%$.
Overall, MLLM-based auditing offers a scalable approach for cross-national youth-safety audits, while provider safety filters ($1.1\%$ refusal rate) under-count the most explicit harms.
\end{abstract}

\section{Introduction}
\label{sec:intro}

TikTok has become one of the most influential gateways through which young users encounter video content online, with a For-You feed that rapidly adapts to micro-interactions such as watch time and loop rates \citep{boeker2022personalization,baumann2025amplification}.
Independent audits have documented self-harm and suicidal-ideation content reaching newly created teen accounts within hours \citep{amnesty2023darkness,amnesty2025rabbithole}, weight-normative and pro-eating-disorder content dominating health-adjacent feeds \citep{minadeo2022weightnormative,strickland2025tiktokdisordered,blackburn2024proana}, and age-gating mechanisms that do not meaningfully shield underage personas relative to adults \citep{eltaher2025protectingyoung,xue2025youthsafety}.
Public-health concerns \citep{murthy2023surgeongeneral} and regulatory frameworks like the EU Digital Services Act make independent, reproducible audits of short-video platforms increasingly urgent.

One response to this moderation burden is to enlist multimodal large language models (MLLMs) as automated annotators: recent work shows aligned hate-speech judgments with humans \citep{davidson2025multimodal,jo2024mllmannotators} and improved alignment from policy- and rule-grounded MLLMs \citep{wu2025icmassistant}.
At the same time, MLLMs disagree substantially with each other on the same items \citep{fasching2025modeldependent} and can exploit language priors to answer without looking at the video \citep{apple2025breakingdown}; how these strengths and failure modes interact with \emph{age} (which shifts the harm distribution the algorithm exposes), \emph{country} (which introduces cross-lingual moderation challenges), and \emph{input modality}, the three dimensions most relevant to youth-safety audits of TikTok, remains largely unexplored.

We address three research questions:
\begin{itemize}
    \item \textbf{RQ1:} Which MLLM configuration agrees with native-speaker annotators well enough to run at scale?
    \item \textbf{RQ2:} What is the marginal value of sampled frames and native video over text-only input?
    \item \textbf{RQ3:} How does harm exposure differ across age personas, countries, and platform signals?
\end{itemize}
We tackle them with a two-stage audit on three EU countries (France, Italy, Sweden) and four age personas (13, 16, 19, 40, spanning TikTok's stated minimum, mid-adolescence, young adulthood in the platform's 18+ tier, and an adult control), via a within-account \texttt{scroll-pre} $\to$ \texttt{SEARCH} $\to$ \texttt{scroll-post} cycle capturing both baseline FYP exposure and the platform's response to an active probe.
Stage-1 selects a validated MLLM auditor on a 300-video annotated subset; Stage-2 runs it on a phase-stratified $10\%$ sample of the full $36{,}971$-video corpus.

On RQ3, within-account harm-keyword search reaches $35$--$56\%$ harm against an immediate \texttt{scroll-pre} baseline of $9$--$44\%$ on the same accounts, and the scroll-post snapshot reverts to baseline; under our Stage-1-validated Gemini 2.5 Flash E3 auditor, Italy is the most exposed country on the three youngest age personas, with its two youngest combinations already at the ceiling at \texttt{scroll-pre}.
The full audit runs at approximately \$49 of API spend.

\paragraph{Our contributions are:}
\begin{itemize}
    \item \textbf{Empirical findings on TikTok harm exposure} across three countries (France, Italy, Sweden) and four age personas (13, 16, 19, 40): keyword search spikes harm exposure $1.5$--$7.5\times$ over passive scrolling, the spike is temporary, age differences flatten under search, and Italy has the highest passive harm rate at every age.
    \item \textbf{An automated harm-annotation pipeline}: we use a multimodal language model (Gemini 2.5 Flash) as the annotator, validated against native-speaker labels on a 300-video subset and then applied to the full corpus for approximately \$50 in API costs.
    \item \textbf{Released artifacts}: a 13-category harm taxonomy aligned with TikTok's Community Guidelines \citep{tiktok2025policies}, the annotation schema and prompts, the per-country keyword lists, and metadata for the $36{,}971$-video corpus.
\end{itemize}

\section{Related Work}
\label{sec:related}

\paragraph{TikTok audits and youth safety}
TikTok has been a recurring target for algorithmic auditing since \citet{boeker2022personalization} isolated the personalization factors that drive its For-You feed, and \citet{baumann2025amplification} subsequently modeled how engagement vectors exponentially amplify niche content.
Qualitative work has documented what this amplification surfaces in practice, particularly weight-normative and pro-eating-disorder content \citep{minadeo2022weightnormative}.
The methodological grounding for the sockpuppet protocol we use comes from \citet{sandvig2014auditing}, who articulated persona-based audits as a rigorous analogue of offline discrimination studies; \citet{le2025autolike} is a recent TikTok-specific instantiation that isolates the effect of interaction signals (e.g., likes) on recommendations.
Closest in design to our work are the two age-stratified TikTok audits: \citet{xue2025youthsafety} simulate age-specific sockpuppets under passive and active engagement and find that platform enforcement does not meaningfully shield under-18 accounts, and \citet{eltaher2025protectingyoung} report a cross-platform version via manual harm annotation, showing that 13-year-old accounts encounter harmful content significantly more often than 18-year-old accounts on TikTok, YouTube, and Instagram.

\paragraph{MLLMs as content moderators and annotators}
\citet{davidson2025multimodal} find that frontier multimodal LLMs can produce context-sensitive hate evaluations that align with aggregate human judgment, supporting the basic feasibility of MLLM-as-auditor use.
\citet{fasching2025modeldependent} demonstrate that different LLM-based moderators disagree substantially on the same items, which is why our Stage-1 compares four model families rather than committing to a single frontier model.
\citet{apple2025breakingdown} show that video LLMs often exploit language priors rather than performing genuine temporal reasoning, which is the failure mode our text-only E1 baseline is designed to surface.
\citet{huang2024legitimacy} reframes the moderation-evaluation problem from accuracy toward legitimacy, the framing required when LLMs are proposed as substitutes for human moderators on platforms with global user bases.

\paragraph{Cross-lingual moderation and frame-based MLLM annotation}
\citet{tonneau2025languagemoderation} analyze EU Digital Services Act transparency reports to quantify content-moderator workforces across languages and platforms; the three countries in our study (France, Italy, Sweden) span the upper, middle, and lower end of TikTok's per-language moderator allocation in that data.
On the prompting side, \citet{wu2025icmassistant} provide evidence that policy- and rule-grounded multimodal LLMs substantially improve alignment with formal community guidelines, motivating the 13-category taxonomy injection in our system prompt (\S\ref{sec:method:models}).
On the input side, \citet{jo2024mllmannotators} use sampled frames plus thumbnail and text metadata to evaluate GPT-4-Turbo against crowdworkers on 19k YouTube videos; their fourteen-frame protocol directly inspired our eight-frame E3 condition (\S\ref{sec:method:models}).

\begin{figure*}[t]
\centering
\includegraphics[width=\textwidth]{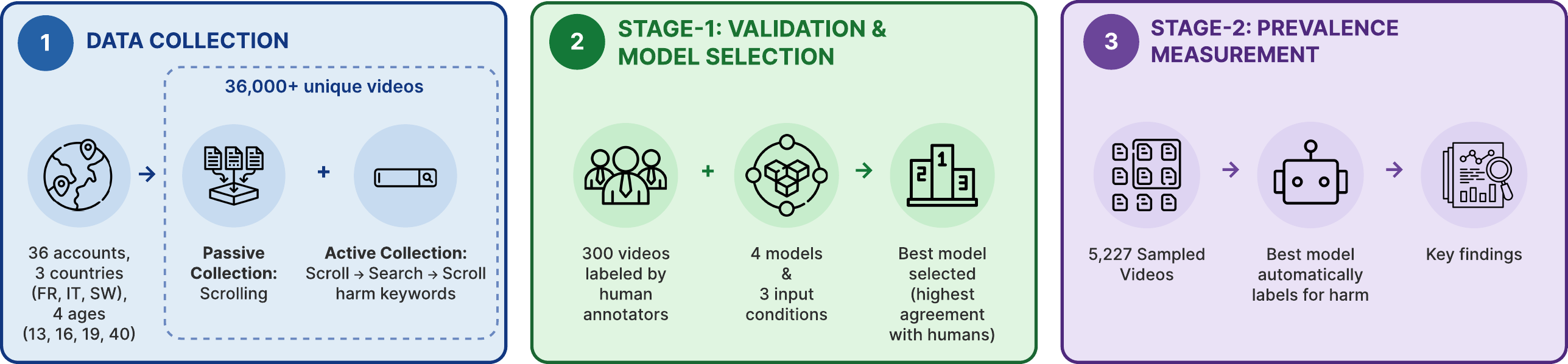}
\caption{End-to-end audit pipeline: data collection, Stage-1 MLLM validation, Stage-2 prevalence measurement.}
\label{fig:pipeline}
\end{figure*}

\section{Methodology}
\label{sec:method}

The audit proceeds in two stages.
Stage-1 selects the MLLM auditor: four candidate models are scored against a 300-video reference set, drawn at random with 25 videos per (country, age) cell, independently labeled by two native-speaker annotators per country and reconciled through joint resolution, and one configuration is carried forward.
The draw is stratified by country and age but not by collection phase; its harm-category mix is not controlled, since categories are only known after annotation.
Stage-2 applies that configuration to a phase-stratified 10\% sample of the full corpus under both native-video upload (E2) and an eight-frame protocol (E3).
All exposure results in \S\ref{sec:results} come from Stage-2; Stage-1 supplies the validation behind them.
Figure~\ref{fig:pipeline} summarizes the end-to-end pipeline.

\subsection{Data Collection}
\label{sec:method:data}

We audit TikTok through persona accounts organized on two axes: country and user age. Each of the three countries in our study (France (FR), Italy (IT), and Sweden (SW)) is paired with four age personas, 13, 16, 19, and 40, yielding twelve (country, age) combinations. For each combination, we use three independent accounts to avoid relying on the behavior of a single account. All accounts report their age during registration and use the system locale in the account's native language. We set the account location by routing traffic through a VPN endpoint in the target country, since TikTok's recommendation and search systems rely on the country linked to the IP address rather than the region listed in the account profile. Without the VPN, feeds tend to match the VPN-free IP location instead of the intended persona location. Each account is driven by a Tampermonkey userscript injected into the TikTok web client that captures the platform's API responses and emulates user scrolling with a randomized scroll amount and an inter-scroll delay drawn uniformly from $500$--$1000$~ms; the three accounts per combination run in parallel on separate browser instances. We use the same personas in both collection phases described below. Sockpuppet-based audits of recommender systems have a long methodological tradition \citep{sandvig2014auditing,boeker2022personalization}, with recent applications to TikTok auditing both age-stratified content exposure \citep{xue2025youthsafety,eltaher2025protectingyoung} and political content skew \citep{ibrahim2026partisan}.

We selected France, Italy, and Sweden because they span a high/medium/low range of TikTok per-language EU moderator allocation in the DSA transparency data analyzed by \citet{tonneau2025languagemoderation} (averaged across 2023--2024 reporting periods: French $\sim$620, Italian $\sim$396, Swedish $\sim$98 moderators).

Data collection proceeds in two phases. The passive phase records what the algorithm surfaces under pure scrolling: each account consumes the For-You feed (FYP) and captures the sequence of recommended videos together with the interactions TikTok exposes in its API responses. The passive phase measures what the algorithm surfaces when no active intent is signaled; this matches prior work showing that engagement-driven amplification is highly sensitive to watch-time patterns even under minimal interaction \citep{boeker2022personalization,baumann2025amplification}. The passive phase was collected between 30 December 2025 and 11 January 2026, five consecutive days per country, and contains 14{,}093 unique videos across the twelve (country, age) combinations (Table~\ref{tab:dataset_overview}).

The active phase probes how the platform responds to keyword-level intent. For each country we curate three native-language keywords per harm category (seven \texttt{SEARCH}-probed categories from \S\ref{sec:method:taxonomy}; full list in Appendix~\ref{app:keywords}). Each active run follows a \texttt{scroll-pre} $\rightarrow$ \texttt{SEARCH} $\rightarrow$ \texttt{scroll-post} cycle. Active collection ran for five days per country between 13 and 19 April 2026 (three weekdays and two weekend days), with three accounts per (country, age) combination scrolling and probing concurrently. The protocol yielded 47{,}674 capture events covering 22{,}878 unique videos (FR: 7{,}130; IT: 7{,}531; SW: 8{,}217). Every capture is tagged with its phase, keyword, category, and originating account.

\paragraph{Dataset overview}
Table~\ref{tab:dataset_overview} reports the unique-video yield per phase, country, and age persona. Together the two phases cover 36{,}971 unique videos (14{,}093 passive + 22{,}878 active). Per-combination counts are balanced to first order across both phases, with each (country, age) combination contributing roughly 1{,}100--2{,}200 unique videos across both phases. Full per-(country, age, phase) breakdowns and engagement statistics are in Appendix~\ref{app:stats}.
English dominates passive feeds in every country (FR 45.7\%, IT 30.3\%, SW 48.3\%), with native-language content at only $10.5$--$20.6\%$.
Native-language keyword queries in the active phase lift native content to $30$--$39\%$ and reduce the English share to $23.9$--$28.9\%$.
Full breakdown in Appendix~\ref{app:stage2_supp}, Table~\ref{tab:lang_overview}.

\begin{table}[t]
\centering
\small
\begin{tabular}{llrr}
\toprule
Country & Age & Passive & Active \\
\midrule
\multirow{5}{*}{FR} & 13           & 1{,}742          & 1{,}685 \\
                    & 16           & 1{,}466          & 1{,}699 \\
                    & 19           & 2{,}175          & 1{,}691 \\
                    & 40           & 1{,}459          & 2{,}055 \\
                    & \textbf{All} & \textbf{6{,}842} & \textbf{7{,}130} \\
\midrule
\multirow{5}{*}{IT} & 13           & 1{,}381          & 1{,}811 \\
                    & 16           & 1{,}355          & 1{,}796 \\
                    & 19           & 1{,}483          & 2{,}022 \\
                    & 40           & 1{,}125          & 1{,}902 \\
                    & \textbf{All} & \textbf{5{,}344} & \textbf{7{,}531} \\
\midrule
\multirow{5}{*}{SW} & 13           & 1{,}565          & 2{,}086 \\
                    & 16           & 1{,}425          & 2{,}150 \\
                    & 19           & 1{,}558          & 1{,}990 \\
                    & 40           & 1{,}374          & 1{,}991 \\
                    & \textbf{All} & \textbf{5{,}922} & \textbf{8{,}217} \\
\midrule
\multicolumn{2}{l}{\textbf{All countries}} & \textbf{14{,}093} & \textbf{22{,}878} \\
\bottomrule
\end{tabular}
\caption{Unique videos per phase, country, and age persona ($36{,}971$ in total; the cross-country overlap of videos collected is $\approx 22\%$ in the passive approach).}
\label{tab:dataset_overview}
\end{table}

\subsection{Harm Taxonomy and Manual Annotation Schema}
\label{sec:method:taxonomy}

\paragraph{Taxonomy}
We adopt a 13-category harm taxonomy aligned with TikTok's Community Guidelines \citep{tiktok2025policies}, spanning disordered eating, self-harm, dangerous challenges, nudity, sexually suggestive content, shocking/graphic content, hate speech, sexual abuse, trafficking, gambling, alcohol/tobacco/drugs, integrity, and harassment (full list with definitions in Appendix~\ref{app:prompts}).
Compared with the six-category taxonomy of \citet{jo2024mllmannotators} and the cross-platform typology of \citet{wef2023typology}, this taxonomy is finer-grained and lets us probe category-level asymmetries that coarser schemes hide.

\paragraph{Annotation schema}
Each video receives a three-way verdict (\textsc{Harmful}, \textsc{Not Harmful}, or \textsc{Video Not Available} when the embed fails or the content is region-locked) and, if \textsc{Harmful}, a required \textit{primary} and optional \textit{secondary} harm category (max two per video). Hesitation is captured by a required {Confidence} rating (\textsc{Low}, \textsc{Medium}, \textsc{High}) rather than a ``borderline'' option, and annotators are instructed to mark \textsc{Low} whenever they hesitate.

\begin{figure}[!t]
\centering
\includesvg[width=0.75\linewidth]{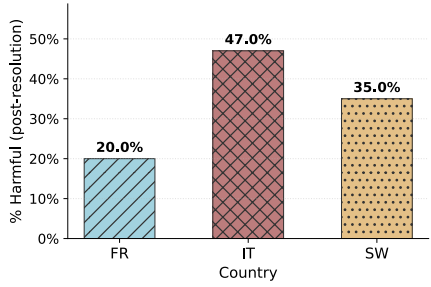}
\caption{Per-country share of \textsc{Harmful} labels in the 300-video Stage-1 reference subset after annotator resolution ($n = 100$ per country). Class balance differs across countries: roughly $1\!:\!4$ in France, $1\!:\!1$ in Italy, $1\!:\!2$ in Sweden.}
\label{fig:stage1_reference_balance}
\end{figure}

\paragraph{Annotators}
Two native-speaker annotators per country independently label the sampled subset and reconcile disagreements through joint resolution to produce \textit{final reference labels} (per-country class balance in Fig.~\ref{fig:stage1_reference_balance}); the resolution rule for subcategory disjunction and pre-resolution inter-annotator agreement are reported in Appendix~\ref{app:stage1_detail}.
The annotators are native-speaker graduate students recruited among the authors' colleagues; they were briefed on the taxonomy, informed that the material could include distressing content, consented to labeling it for research, and worked in self-paced sessions through a dedicated web panel.
They were not financially compensated, and each annotator spent approximately five days on the task.

\subsection{Models and Experimental Conditions}
\label{sec:method:models}

\paragraph{Models}
We evaluate four MLLMs spanning three provider families: Gemini 2.5 Flash (Google AI SDK, native-video capable), Qwen3-VL-32B (Alibaba DashScope, native-video capable), GPT-4o-mini, and Mistral Large 3 (both via OpenRouter).
Two models process video natively at the API layer; the other two ingest only text and discrete frames.
Since different MLLMs disagree substantially on the same moderation items \citep{fasching2025modeldependent}, we read aggregate performance and pairwise disagreement side by side.

\paragraph{Three input conditions}
Each model is evaluated under three conditions: E1 (text-only), the video caption plus an audio transcript, isolating the linguistic prior \citep{apple2025breakingdown}; E2 (native video), the raw MP4 uploaded through the provider's video API, applicable only to the two native-video-capable models; and E3 (frames-as-images), eight frames sampled uniformly over the video and submitted as inline base64 images alongside the caption and transcript, following the frame-plus-metadata protocol of \citet{jo2024mllmannotators}.

\paragraph{Prompting}
All conditions share a system prompt that injects the 13-category taxonomy with one-sentence definitions, following \citet{wu2025icmassistant}'s evidence that policy- and rule-grounded MLLM moderation aligns better with formal guidelines.
The model returns structured JSON with the verdict, a primary harm subcategory if \textit{harmful}, and a free-text reasoning span; the confidence rating is collected from human annotators only.
Full prompt text is in Appendix~\ref{app:prompts}.

\subsection{Stage-2 Sample Selection}
\label{sec:method:stage2}

Stage~2 applies the Stage~1 winning auditor, Gemini 2.5 Flash, to a stratified sample of the $36{,}971$-video corpus.
We first draw $10\%$ of videos separately within each country, age persona, and phase, where the phases are passive For-You-page scrolling and the three active-session steps: before search, search results, and after search.
Because the initial draw contained too few before-search and after-search videos for a stable search-versus-baseline comparison, we top up these two active sub-phases from the same accounts and collection window until each country--age combination contains roughly $100$ videos per phase.
The resulting Stage-2 sample contains $1{,}817$ passive videos and $4{,}251$ active-session videos; after removing unavailable videos, provider refusals, and parse failures, the final sample yields $5{,}227$ usable E3 verdicts and $5{,}248$ usable E2 verdicts, with $5{,}108$ paired videos.

Because the passive and active collections were gathered in separate time windows, we analyze them separately throughout \S\ref{sec:results}.
The main findings also hold on the original $10\%$ draw.
We omit E1 at Stage~2 because text-only input performed poorly in Stage~1.

\section{Results}
\label{sec:results}

\subsection{MLLM Validation (Stage-1)}
\label{sec:results:agreement}

\begin{figure*}[!t]
\centering
\includesvg[width=\textwidth]{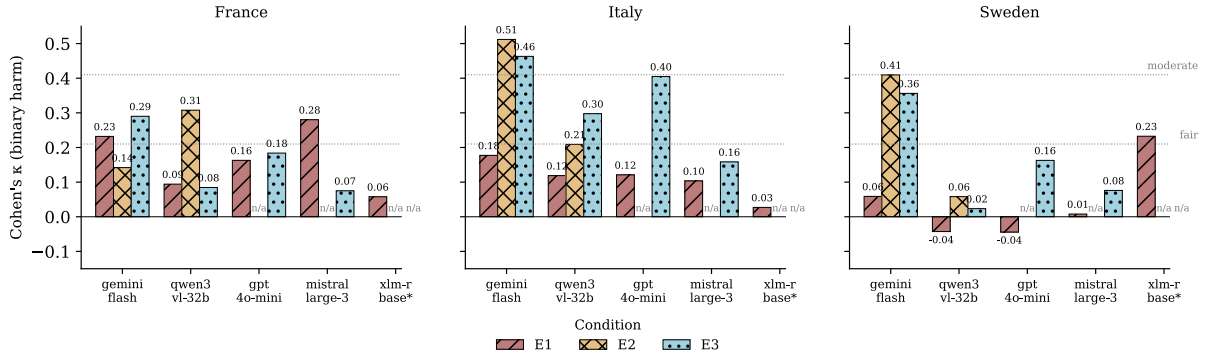}
\caption{Per-country Cohen's $\kappa$ on the binary harm verdict for the ten Stage-1 (model, condition) combinations (E1: text-only; E2: native video; E3: eight frames plus text). Dotted lines: $\kappa = 0.21$ and $\kappa = 0.41$ \citep{landis1977kappa}.}
\label{fig:stage1_kappa_grid_by_country}
\end{figure*}

We first evaluate the accuracy of different models at identifying harmful vs.\ non-harmful content.
Across the ten Stage-1 (model, condition) combinations (Fig.~\ref{fig:stage1_kappa_grid_by_country}), Gemini 2.5 Flash under the eight-frame condition (E3) is the strongest configuration overall, but even it tops out at aggregate Cohen's $\kappa = 0.42$ and no (model, condition) clears moderate agreement \citep{landis1977kappa} in every country, so the task is hard.
For Gemini~E3, Italian content reaches moderate agreement while French and Swedish stay at fair agreement; the cross-country $\kappa$ ranking tracks the cross-country class-balance ranking in the reference subset (Fig.~\ref{fig:stage1_reference_balance}), which mechanically suppresses $\kappa$ on the most-imbalanced country.
Within Gemini specifically, agreement improves as more visual input is added (E1 $\kappa = 0.17 \to$ E2 $\kappa = 0.38 \to$ E3 $\kappa = 0.42$; Appendix~\ref{app:stage1_detail}, Figure~\ref{fig:stage1_modality_monotonicity}), and the eight-frame condition outperforms native-video upload at approximately $2\times$ lower per-call cost.
All non-Gemini configurations sit below $\kappa = 0.30$ in the aggregate, but the per-country profile is uneven: GPT-4o-mini's eight-frame run is the strongest non-Gemini combination at aggregate $\kappa = 0.29$ and reaches moderate agreement on Italian content, so the model ranking re-orders by country.
We therefore carry Gemini's two top configurations (E3 as primary, E2 alongside for robustness) into Stage-2 and drop the text-only E1 pathway, since no model clears fair agreement on text alone.
The remaining diagnostic plots (aggregate grid, inter-model agreement, confusion flow, per-country category mix) and the failure-counts table are in Appendix~\ref{app:stage1_detail}.

E2 reports harm rates $3$--$12$~pp higher than E3 across the same combinations (e.g.\ FR-16: $35.1\%$ vs.\ $28.1\%$; SW-13: $31.3\%$ vs.\ $24.2\%$).
The two modalities agree on the binary harm verdict at $\kappa = 0.614$ over $n = 5{,}108$ paired items (Table~\ref{tab:stage2_e2e3}), with per-country $\kappa \in [0.60, 0.63]$, so the cross-combination ordering is preserved.
We treat E3 as the primary reference because Stage-1 validated it against the annotator; E2's higher reporting rate is consistent with a more permissive judgment layer when the full clip is available, but at this scale we cannot resolve it without a Stage-2 annotation pass.
The E2-vs-E3 disagreement is not uniform across harm categories: E2 over-flags visual-cue categories (Sexually Suggestive, Shocking and Graphic, Nudity) while E3 picks up more dialogic Harassment and Bullying (per-category decomposition in Appendix~\ref{app:stage2_supp}, Table~\ref{tab:stage2_e2e3_per_cat}).

\subsection{Cross-Country, Cross-Age Harm Prevalence}
\label{sec:results:prevalence}

All rates in \S\ref{sec:results:prevalence}--\S\ref{sec:results:phase} are estimates under the Stage-1-validated Gemini~E3 auditor, and denominators include every sampled video served to the persona, regardless of detected content language.
\textbf{Italy carries the highest estimated harm rate across all four age personas}, with IT-16 the most-exposed case in the audit at $40.4\%$.
France ranges from $23.6\%$ to $32.4\%$; Sweden from $24.2\%$ to $31.0\%$.
A per-country precision/recall correction derived from Stage-1 (full derivation in Appendix~\ref{app:stage1_detail}) leaves this ordering intact at ages 13, 16, and 19, but at age 40 the corrected French rate ($\approx 49\%$) overtakes the corrected Italian ($\approx 38\%$) and Swedish ($\approx 33\%$) rates, so we read the age-40 comparison with more caution than the three younger ages.
Within each country, harm rates are relatively flat across the four age personas (a $7$--$9$~pp spread in France and Sweden); the exception is Italy, where the $13$-year-old persona sees more harm than the adult, inverting the youth-safety-first prior.
E2 (native video) reports systematically higher harm rates than E3 (eight frames plus text) at every (country, age) combination (per-combination breakdown with E2 alongside in Appendix~\ref{app:stage2_supp}, Figure~\ref{fig:stage2_harm_country_age}; modality comparison in \S\ref{sec:results:agreement}).

\begin{figure}[!t]
\centering
\includesvg[width=\linewidth]{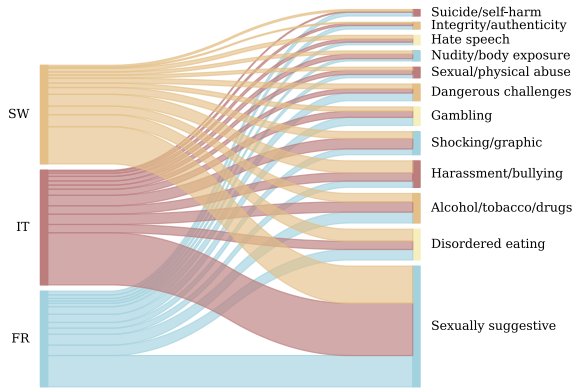}
\caption{Country~$\to$~harm-subcategory aggregate Sankey over Gemini~E3 flagged items at Stage-2. Ribbon widths are proportional to absolute counts.}
\label{fig:stage2_subcategory_sankey}
\end{figure}

Sexually suggestive content is the dominant harm category in every (country, age) combination, from $21.7\%$ of harmful items (SW-13) to $50.0\%$ (IT-16); the per-country breakdown of E3-flagged items by harm subcategory is shown in Figure~\ref{fig:stage2_subcategory_sankey}, with the caveat that Stage-1 strict primary-subcategory agreement is only $29\%$ ($54\%$ under primary-or-secondary matching; Appendix~\ref{app:stage1_detail}).

\subsection{Passive-Phase Exposure}
\label{sec:results:passive}

\begin{figure}[t]
\centering
\includesvg[width=\linewidth]{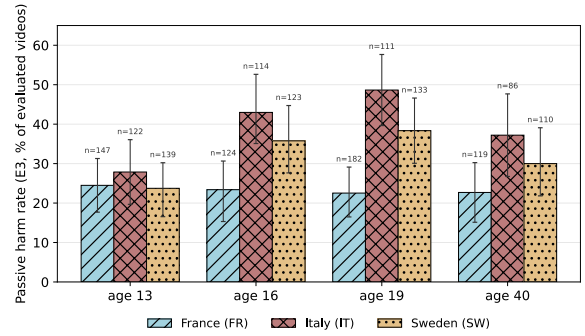}
\caption{Stage-2 harm rate per (country, age) on the passive collection phase (FYP scrolling only, no search probe) under Gemini~E3, with $95\%$ video-level bootstrap CIs. Italy leads at every age and the cross-country gap widens with age. E2 version in Appendix~\ref{app:stage2_supp}, Figure~\ref{fig:stage2_harm_passive_e2}.}
\label{fig:stage2_harm_passive_e3}
\end{figure}

Under passive FYP scrolling (no search probe), Italy carries the highest harm rate at every age (Fig.~\ref{fig:stage2_harm_passive_e3}), with the cross-country gap widening from a $4$-point spread at age 13 (FR $24.5\%$, IT $27.9\%$, SW $23.7\%$) to over $25$ points at age 19 (FR $22.5\%$, IT $48.6\%$, SW $38.4\%$); Italian age-19 is the most-exposed case at $48.6\%$.
France stays flat at ${\sim}23\%$ across all four ages, while Sweden rises from $24\%$ (age 13) to $38\%$ (age 19).
The E2 modality reports a few percentage points higher in every combination but preserves the same cross-country ordering (Appendix~\ref{app:stage2_supp}, Figure~\ref{fig:stage2_harm_passive_e2}).
Because the passive phase and the active phase were collected in separate windows, we analyze them separately rather than as parallel signals at one point in time (\S\ref{sec:results:phase}); Italy's lead holds in both analyses.

\subsection{Search-Phase Exposure}
\label{sec:results:phase}

\begin{figure}[!t]
\centering
\includesvg[width=1.0\linewidth]{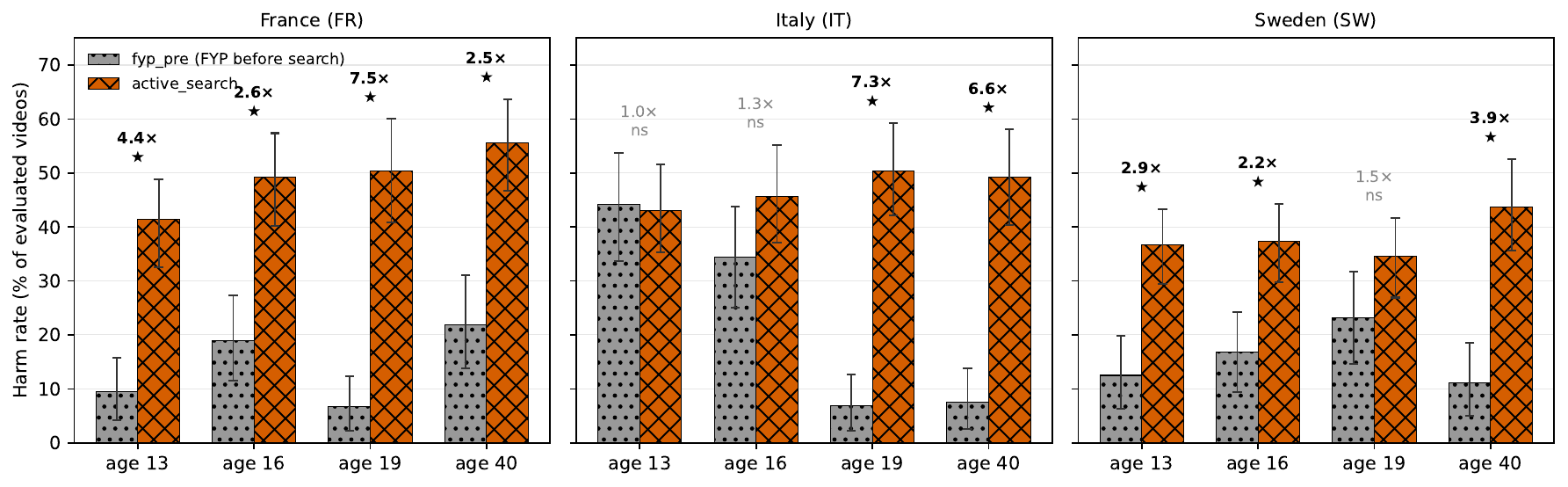}
\caption{Per-combination harm rate at the three steps of the active collection phase (\texttt{scroll-pre} $\to$ \texttt{SEARCH} keyword probe $\to$ \texttt{scroll-post}) on E3 with \% CIs. Annotated: \texttt{SEARCH}/\texttt{scroll-pre} ratio; stars mark non-overlapping CIs.}
\label{fig:stage2_amplification}
\end{figure}

The within-account \texttt{scroll-pre} $\to$ \texttt{SEARCH} $\to$ \texttt{scroll-post} cycle measures how the algorithm responds to a harm-keyword search and whether the probe contaminates the scroll-post FYP (Fig.~\ref{fig:stage2_amplification}).
{The \texttt{SEARCH} endpoint returns $35$--$56\%$ harmful content ($1.5$--$7.5\times$ over \texttt{scroll-pre} in ten of twelve combinations) despite no visible search-time block or warning in our audit, while \texttt{scroll-post} reverts to within a few percentage points of \texttt{scroll-pre}.}
A higher harm rate under harm-keyword search than under scrolling is expected by construction; the findings are its magnitude, the collapse of the age gradient, and the absence of visible search-time intervention, not evidence of a general moderation failure.

The largest lifts are on older French and Italian personas, whose \texttt{scroll-pre} rate sits below $25\%$ and whose \texttt{SEARCH} endpoint returns $50$--$56\%$ harmful items before falling back; Sweden's adult combination shows the same pattern at smaller magnitude.
The two exceptions are IT-13 and IT-16, where \texttt{scroll-pre} is already at $34$--$44\%$ harm: the harm-keyword search has no headroom to lift further, and \texttt{scroll-post} stays in range of \texttt{scroll-pre}.
The contrast with the same-age French and Swedish combinations, which sit at $9$--$19\%$ pre-probe, suggests that country and not persona age sets the headroom on this audit window.
A related observation is that, at the \texttt{SEARCH} endpoint, harm rates collapse into a $35$--$56\%$ band across all twelve combinations: the youngest personas in France and Sweden reach $37$--$42\%$ harm under search, within a few percentage points of the adult combinations in the same countries, so the \texttt{scroll-pre} baseline age gradient is wiped out by the probe.
A second pattern is the speed of the reversion.
Across all twelve combinations, the \texttt{scroll-pre} and \texttt{scroll-post} \% CIs overlap and the absolute change is bounded by a few percentage points.
Within the five-day collection window the harm-keyword search elevates exposure during the search session itself, not as a persistent recommendation-feed drift.

\paragraph{Account-clustered uncertainty}
Because each cell is instantiated by three accounts and recommendations are sequential, a video-level bootstrap could understate uncertainty.
Recomputing every active-phase interval with a hierarchical bootstrap that resamples accounts before videos leaves the intervals essentially unchanged (median width ratio $1.00$, at most $1.85\times$ on the smallest \texttt{scroll-post} cells), and per-account \texttt{SEARCH} harm rates within a cell differ by at most a few percentage points (Appendix~\ref{app:stage2_supp}, Table~\ref{tab:clustered_ci}).
The clustered intervals separate \texttt{SEARCH} from \texttt{scroll-pre} in ten of twelve combinations (all but the two Italian ceiling cells IT-13 and IT-16) and preserve the \texttt{scroll-pre}/\texttt{scroll-post} overlap in all twelve.
Per-account attribution is available for the active phase only, so passive-phase CIs (\S\ref{sec:results:passive}) remain video-level.

\paragraph{Policy-tier split of the age gradient}
The binary verdict pools content TikTok prohibits for every audience with content it permits for adults but restricts for minors.
Splitting the 13 categories into an 18+-restricted tier (Sexually Suggestive, Nudity, Alcohol/Tobacco/Drugs, Gambling, Shocking and Graphic) and a universally prohibited tier separates the two readings (Appendix~\ref{app:stage2_supp}, Table~\ref{tab:age_policy_split}).
Under passive scrolling the restricted tier rises with persona age in Italy and Sweden (IT $17.2$~pp at age 13 vs.\ $26.7$~pp at 40; SW $8.6$ vs.\ $24.5$~pp), the direction age gating predicts, while the prohibited tier does not fall for minors (SW-13 carries the audit's highest passive prohibited-tier rate at $15.1$~pp).
Under \texttt{SEARCH} the age separation on the restricted tier disappears: the 13-year-old personas receive $24$--$35$~pp of 18+-restricted content, within a few points of the adult personas in the same countries.
We do not evaluate compliance with TikTok's age policies as such; the split shows that the flat total-rate age gradient mixes a rising restricted tier with a non-declining prohibited tier rather than indicating uniform age-blindness.

\paragraph{Temporal stability of the audit window}
A day-by-day breakdown of Gemini~E3 harm rate per (country, age, signal) on the active sample (Appendix~\ref{app:stage2_supp}, Figure~\ref{fig:stage2_daily_active}) shows the \texttt{SEARCH} lift and the Italian \texttt{scroll-pre} ceiling effect hold on every day of the five-day window; the cross-combination averages in \S\ref{sec:results:prevalence}--\S\ref{sec:results:phase} are not artifacts of a single-day spike.

\subsection{Provider Blocks and Failure Modes at Scale}
\label{sec:results:failures}

The Gemini safety layer refuses to score about $1.1\%$ of Stage-2 inputs (per-country breakdown in Appendix~\ref{app:stage2_supp}, Table~\ref{tab:stage2_blocks_summary}); a small additional set of empty responses and transient network errors accounts for the gap between the ${\sim}5{,}300$ inputs per modality and the headline denominators.
The block rate is broadly balanced across countries and comparable across modalities (E2: $57/{\sim}5{,}300 \approx 1.08\%$; E3: $64/{\sim}5{,}300 \approx 1.21\%$; Fisher exact $p \approx 0.55$).
The blocks are not noise-distributed across categories. Only $23$ of the $64$ E3 refusals are surfaced through a \texttt{SEARCH} keyword and therefore category-attributable; the remaining $41$ sit in \texttt{scroll-pre}, \texttt{scroll-post}, or passive-only paths, where no keyword fixes a harm category. Among the $23$ \texttt{SEARCH}-attributable blocks, the per-category block rate is dominated by Nudity and Body Exposure ($\approx 4.8\%$), with Sexually Suggestive Content ($\approx 2.6\%$), Shocking and Graphic Content ($\approx 0.7\%$), and Disordered Eating and Body Image ($\approx 0.5\%$) next; the remaining three search-targeted categories (Dangerous Challenges, Gambling, Alcohol/Tobacco/Drugs) yielded zero or near-zero blocks.
Reported prevalences are therefore under-estimates for exactly the categories the audit is designed to surface, with the largest measurement bias on Nudity and Suggestive content.
A worst-case sensitivity bound (assigning every block-or-parse-fail item to either \textsc{Harmful} or \textsc{Not Harmful}) shifts headline rates by at most $1$--$3$~pp and preserves both the cross-country IT~$>$~SW~$\geq$~FR ordering and the IT-13/16 ceiling-effect finding (Appendix~\ref{app:stage2_supp}).

\section{Discussion and Conclusion}
\label{sec:discussion}

\paragraph{MLLM auditing is feasible at this scale, conditional on three structural caveats}
First, the Stage-1 $\kappa = 0.42$ was measured on the 300-video reference sample, and whether it transfers to the full Stage-2 distribution is an open question this design cannot answer.
Second, at scale E2 (native video) reports consistently higher harm rates than primary E3 (eight frames plus text) ($+3$--$12$ pp), and which modality is closer to human truth on the Stage-2 population is not answerable without a Stage-2 annotation pass.
Third, the provider safety layer refuses $1.1\%$ of inputs non-randomly with respect to harm category (\S\ref{sec:results:failures}), so reported prevalences are under-estimates for the most explicit content.
MLLM auditing is therefore cheap enough at full corpus scale under these assumptions, but not yet a drop-in replacement for human annotation on policy-edge items.

\paragraph{Cross-country variation is large, and the patterns differ}
Under E3, Italy carries the highest measured harm rate across all four age personas, with the age-40 case shifting under per-country precision/recall recalibration (Appendix~\ref{app:stage1_detail}) but the strongest reading holding on the three youngest, and the within-country age gradients run in opposite directions across the three countries.
The same Italy-leads pattern replicates under purely passive FYP scrolling (\S\ref{sec:results:passive}), with Italian age-19 reaching $48.6\%$ harm, the highest measured rate in the audit, against ${\sim}23\%$ in France at the same age.
The two youngest Italian combinations (IT-13, IT-16) are already at $34$--$44\%$ harm at \texttt{scroll-pre}, while the same age combinations in France and Sweden sit far lower and respond strongly to the probe.
Three decompositions narrow the candidate explanations for the Italian pattern (Appendix~\ref{app:stage2_supp}).
The lead is concentrated in one category: Sexually Suggestive content contributes $23.8$~pp of Italy's $39.0\%$ passive rate, against $12.2$~pp in France and $15.8$~pp in Sweden.
It is not carried by globally circulating videos: on passive items that also appear in another country's corpus, Italy's estimated rate is $19.2\%$, indistinguishable from France ($26.4\%$) and Sweden ($28.3\%$), while Italy-exclusive content sits at $41.7\%$.
It also survives a language control: English-language videos served to Italian accounts are flagged at $32.4\%$, clearly above the English-language rate in France ($19.5\%$), with Sweden in between ($27.5\%$).
The pattern therefore points to the country-specific slice of the pool TikTok serves to Italian accounts rather than to translation artifacts or annotator thresholds; whether that slice reflects content supply or per-language moderation capacity \citep{tonneau2025languagemoderation} is not identifiable from the outside.

\paragraph{The harm-keyword search endpoint is the main driver of exposure, and the elevated exposure is short-lived}
The $1.5$--$7.5\times$ within-account \texttt{scroll-pre} $\to$ \texttt{SEARCH} lift, paired with near-complete \texttt{scroll-post} reversion (\S\ref{sec:results:phase}), means a passive-FYP-only audit understates the harm a determined user can reach by keyword by an order of magnitude in many combinations; this is search returning what the keyword asks for in the absence of visible search-time moderation, not recommender-side amplification in the sense of \citet{baumann2025amplification}.
The FR/SW scroll-pre age gradient is absent at \texttt{SEARCH}, but our design does not distinguish (a) age-insensitive platform retrieval from (b) a keyword set whose returned content pool overlaps across personas; the same 21 keywords are issued by every persona, and an item-overlap analysis within (country, keyword) would discriminate between these two readings.

\paragraph{Conclusion}
Two findings stand out: keyword search, not recommender amplification, is the dominant harm-exposure pathway in this audit, returning $35$--$56\%$ harmful content for harm-seeking queries with no visible search-time friction, while baseline algorithmic exposure remains sharply uneven across countries, with accounts registered in Italy the most exposed at every age.
Methodologically, MLLM-based auditing scales cross-national youth-safety audits at API costs of order \$50, with annotation throughput rather than compute cost as the remaining scaling bottleneck.
These results point to platform-side search-time moderation and per-language moderation infrastructure as the natural next levers for reducing harm exposure on short-video platforms.

\section{Limitations}
\label{sec:limitations}

Two native-speaker annotators per country produced the Stage-1 final reference labels via joint resolution; these are not statistical ground truth (pre-resolution inter-annotator agreement in Appendix~\ref{app:stage1_detail}; \citealp{artstein2008iaa, krippendorff2011alpha}).
The annotators apply the public text of the Community Guidelines and are not professional content moderators; TikTok's internal enforcement thresholds are not observable, so the reference labels are guideline-grounded judgments rather than platform enforcement truth.

Stage-2 runs a single auditor model with no per-video annotation pass; the E2-vs-E3 cross-modality agreement is an internal-consistency check rather than a second validation against reference labels. Reported Stage-2 harm rates should be read as Gemini~E3-as-auditor estimates whose transfer from the 300-video Stage-1 reference to the full Stage-2 distribution is unverified; the reference draw is stratified by country and age but not by phase, so phase-dependent auditor error cannot be ruled out.

Provider-side refusals (${\sim}1.1\%$ at Stage-2, plus Stage-1 Qwen/DashScope failures in Appendix~\ref{app:stage1_detail}) cluster on the most-explicit harm categories; the worst-case sensitivity bound is in \S\ref{sec:results:failures}.

Our data collection spans a single, time-bounded window, and TikTok's recommender adapts continuously, so exposure patterns measured here may not generalize to other time periods or to persona profiles we did not instantiate. The personas themselves are programmatically controlled accounts that lack the behavioral richness of real users (no multi-device use, no cross-session continuity beyond what we script, no organic social graph), a standard caveat of sockpuppet auditing \citep{sandvig2014auditing}. Findings should therefore be read as upper-bound claims about what the algorithm can serve under simple engagement rules, not as point estimates of real-user exposure. The VPN-routed sockpuppet accounts may also have been handled by TikTok's anti-abuse layer differently from native-IP traffic in ways we did not separately measure; the absence of mass capture failures suggests this was not severe but does not bound the residual effect.

\paragraph{Ethics and data release}
All collected videos are from public TikTok accounts and no real users are impersonated.
Following the platform's terms of service, we plan to release: per-video metadata for the full $36{,}971$-video corpus (video IDs, country, age persona, phase, keyword, capture timestamp, public engagement counts), the per-country keyword lists (Appendix~\ref{app:keywords}), the per-experiment prompt templates (Appendix~\ref{app:prompts}), the Gemini~E3 and E2 verdicts and reasoning spans for the Stage-2 subset, and the aggregated statistics reported in this paper. We do \emph{not} redistribute raw video content, raw API payloads, or any information identifying public TikTok users beyond the video ID; videos that have since been deleted, made private, or geo-restricted are released as IDs only. The Stage-1 annotations are released in aggregate (per-combination agreement summaries) rather than per-video, to avoid re-identifying the two annotators per country.
The collection, evaluation, and analysis code (scrapers, MLLM-evaluation scripts, plotting and statistics pipeline) is released alongside the data. AI assistants were used for writing improvements and editing the text.

\bibliography{custom}

\appendix

\section{Keyword Lists}
\label{app:keywords}

The active-phase \texttt{SEARCH} probe uses three native-language keywords per country and per harm category, listed in Table~\ref{tab:keywords}. Categories without a row were not probed (the remaining six taxonomy categories from \S\ref{sec:method:taxonomy} are not target-able by short keyword queries with reasonable precision). The same English category labels are used as the standard harm-category names throughout the paper; the harm-keyword search issues each keyword in the persona country's native language.

\paragraph{Keyword selection procedure}
The English seed terms for each of the seven probed harm categories were drafted directly from the taxonomy definitions in \S\ref{sec:method:taxonomy}, with three queries per category covering one literal severity-axis anchor (e.g.\ \textit{blood}, \textit{nudity}, \textit{casino}), one moderate variant (e.g.\ \textit{fights}, \textit{body exposure}, \textit{betting}), and one platform-vernacular phrase (e.g.\ \textit{natural disaster}, \textit{intimate kisses}, \textit{sports betting}) chosen to surface content that TikTok's search endpoint actually returns for the category rather than zero-result moderation-jargon strings. Each English triple was then translated by a native-speaker researcher in Italian, French, and Swedish, with the constraint that the per-country translation should preserve the same severity-axis structure (anchor / moderate / vernacular) rather than the literal lexical form, so that the cross-country \texttt{SEARCH}/\texttt{scroll-pre} ratio of \S\ref{sec:results:phase} is read against queries of comparable retrieval intent. The full keyword list was \emph{fixed before any active-phase data collection began} and is reported here unaltered; no post-hoc keyword tuning was applied after observing the \texttt{SEARCH} harm rates. The textual coverage check in Appendix~\ref{app:stage2_supp} (Figure~\ref{fig:stage2_keyword_match}) confirms that the chosen keywords surface keyword-bearing content at a $20$--$25\%$ rate under \texttt{SEARCH} against a $0$--$2\%$ FYP baseline.

\begin{table*}[t]
\centering
\small
\setlength{\tabcolsep}{4pt}
\begin{tabular}{p{3.4cm}p{3.7cm}p{3.7cm}p{3.7cm}}
\toprule
Harm category & French (FR) & Italian (IT) & Swedish (SW) \\
\midrule
Disordered Eating and Body Image &
  régimes restrictifs; perdre du poids; chirurgie esthétique &
  diete restrittive; perdere peso; chirurgia estetica &
  restriktiva dieter; gå ner i vikt; estetisk kirurgi \\
Dangerous Activity and Challenges &
  challenges dangereuses; activités dangereuses; tendance &
  challenge pericolose; attività pericolose; tendenza &
  farliga challenges; farliga aktiviteter; trend \\
Nudity and Body Exposure &
  nudité; exposition corporelle; sous-vêtements &
  nudità; esposizione corporea; biancheria intima &
  nakenhet; kroppsexponering; underkläder \\
Sexually Suggestive Content &
  baisers intimes; sexe; produits sexuels &
  baci intimi; sesso; prodotti per il sesso &
  intima kyssar; sex; sexleksaker \\
Shocking and Graphic Content &
  sang; combats; catastrophe naturelle &
  sangue; combattimenti; disastro naturale &
  blod; slagsmål; naturkatastrof \\
Gambling &
  jeux d'argent; casino; paris sportifs &
  gioco; casinò; scommesse &
  spel; kasino; vadslagning \\
Alcohol, Tobacco, and Drugs &
  drogues; tabac; alcool &
  droghe; tabacco; alcol &
  droger; tobak; alkohol \\
\bottomrule
\end{tabular}
\caption{Per-country, per-category native-language keywords used in the active-phase \texttt{SEARCH} probe (\S\ref{sec:method:data}). Each persona account issues all 21 keywords in its country's language across the five-day collection window; the vocabulary was fixed before data collection.}
\label{tab:keywords}
\end{table*}

\section{Prompt Templates}
\label{app:prompts}

This appendix gives the full prompts used at both stages. The same system prompt is shared across all four MLLMs and across the three input conditions; only the user-message structure changes by condition.

\paragraph{System prompt (all conditions, all models)}

\begin{quote}\small\ttfamily
Classify the following TikTok video as `harmful' or `not harmful' under this harmful content framework (aligned with TikTok Community Guidelines):
\begin{itemize}\setlength{\itemsep}{0pt}
\item \textbf{Disordered Eating and Body Image}: Content that promotes disordered eating, risky weight loss or muscle gain methods, or harmful body comparisons.
\item \textbf{Suicide and Self-Harm}: Content that shows, promotes, or provides instructions for suicide or self-harm.
\item \textbf{Dangerous Activity and Challenges}: Content that shows or promotes dangerous stunts, dares, or challenges that could lead to physical harm.
\item \textbf{Nudity and Body Exposure}: Nudity, sexual activity, or significant body exposure; any such content involving minors is prohibited.
\item \textbf{Sexually Suggestive Content}: Sexually suggestive acts or behavior, or sexually explicit language; any such content involving minors is not allowed.
\item \textbf{Shocking and Graphic Content}: Extremely graphic, violent, or disturbing content that could cause viewers emotional distress.
\item \textbf{Hate Speech or Behavior}: Content that promotes hate or attacks people based on protected attributes.
\item \textbf{Sexual and Physical Abuse}: Content that shows, promotes, or facilitates sexual abuse, exploitation, or physical harm of people (including youth).
\item \textbf{Human Trafficking and Smuggling}: Content that promotes or facilitates human trafficking or smuggling.
\item \textbf{Gambling}: Content that promotes or facilitates gambling or gambling-like activities (e.g.\ social casinos, mystery boxes).
\item \textbf{Alcohol, Tobacco, and Drugs}: Content that trades, markets, or promotes alcohol, tobacco, vaping, e-cigarettes, or illegal or regulated drugs.
\item \textbf{Integrity and Authenticity}: Deceptive behavior or fake engagement; misleading edited or AI-generated content; or misinformation that could cause significant harm.
\item \textbf{Harassment and Bullying}: Content that harasses or bullies others, including degrading remarks about appearance, doxing, sexual harassment, or coordinated abuse.
\end{itemize}

Watch the video carefully (visual content, audio, on-screen text). Also consider the description and, if provided, the comments. The goal is to prevent minors from seeing harmful content.

Reply with a single JSON object and nothing else:
\begin{verbatim}
{
  "verdict": "harmful" or "not harmful",
  "subcategory": "<one of the 13 categories above,
                  or null if not harmful>",
  "reasoning": "<your brief reasoning>"
}
\end{verbatim}

If the video falls under one or more of the categories above in a way that violates the definitions, choose ``harmful''. Otherwise choose ``not harmful''.
\end{quote}

\paragraph{User message per condition}

The user message contains the video content (in the form appropriate for each condition) plus a short text block. The text block is:

\begin{itemize}\setlength{\itemsep}{1pt}
\item \textbf{E1 (text-only):} ``Video description: \{description\}'' + ``Audio transcript:\textbackslash n\{transcript\}'' + ``Classify this video according to the instructions.'' No visual content.
\item \textbf{E2 (native video):} the raw MP4 bytes uploaded through the provider's video API, plus ``Note: no description is provided -- evaluate the video content only. Classify this video according to the instructions.''
\item \textbf{E3 (frames + text):} ``Video description: \{description\}'' + ``The following 8 image(s) are frames extracted at equal intervals from a TikTok video. Analyze them together as a representation of the full video content.'' + the eight frames as inline base64 PNGs + ``Audio transcript:\textbackslash n\{transcript\}'' + ``Classify this video according to the instructions.''
\end{itemize}

The same prompt template is used at Stage-1 (four-model comparison) and Stage-2 (Gemini~E3 run) so that the Stage-1 $\kappa$ calibration transfers without prompt-induced drift.

\section{Dataset Statistics}
\label{app:stats}

This appendix provides the detailed dataset breakdowns summarized in \S\ref{sec:method:data}.


\begin{table*}[ht]
\centering
\small
\begin{tabular}{llrrrrrr}
\toprule
Country & Age & \texttt{scroll-pre} items & \texttt{SEARCH} items & \texttt{scroll-post} items & Unique videos & Mean plays & Mean likes \\
\midrule
FR & 13 & 170 & 3{,}055 & 195 & 1{,}685 & 751{,}k & 28{,}k \\
FR & 16 & 210 & 3{,}038 & 197 & 1{,}699 & 907{,}k & 47{,}k \\
FR & 19 & 217 & 2{,}980 & 203 & 1{,}691 & 856{,}k & 37{,}k \\
FR & 40 & 200 & 3{,}533 & 218 & 2{,}055 & 972{,}k & 42{,}k \\
IT & 13 & 230 & 3{,}583 & 266 & 1{,}811 & 723{,}k & 32{,}k \\
IT & 16 & 254 & 3{,}727 & 228 & 1{,}796 & 890{,}k & 39{,}k \\
IT & 19 & 228 & 4{,}074 & 247 & 2{,}022 & 905{,}k & 31{,}k \\
IT & 40 & 232 & 3{,}728 & 260 & 1{,}902 & 1{,}083{,}k & 37{,}k \\
SW & 13 & 257 & 3{,}570 & 246 & 2{,}086 & 573{,}k & 45{,}k \\
SW & 16 & 253 & 3{,}540 & 252 & 2{,}150 & 620{,}k & 64{,}k \\
SW & 19 & 256 & 3{,}652 & 243 & 1{,}990 & 620{,}k & 45{,}k \\
SW & 40 & 253 & 3{,}637 & 242 & 1{,}991 & 749{,}k & 18{,}k \\
\bottomrule
\end{tabular}
\caption{Per-(country, age) capture counts by phase and unique-video yield, with mean plays and likes over all items in the combination.}
\label{tab:dataset_detail}
\end{table*}

\begin{figure*}[t]
\centering
\includesvg[width=0.95\textwidth]{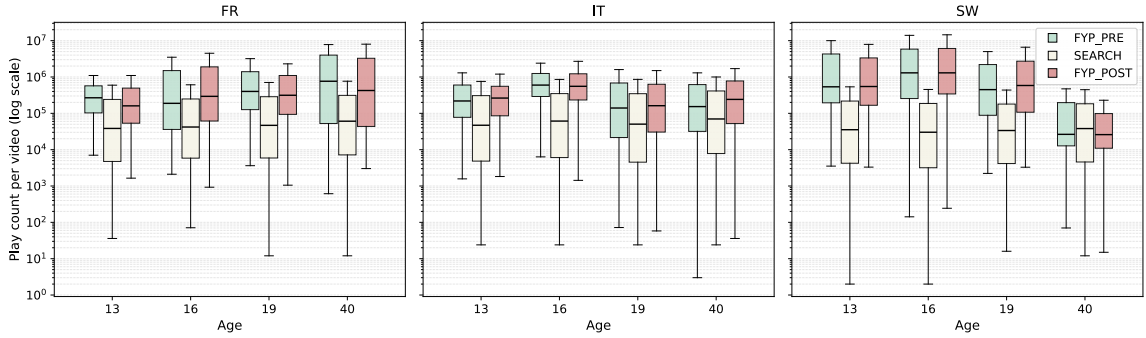}
\caption{Distribution of play counts per video by (country, age, phase) in the active dataset (log scale, one playcount per unique video, outliers suppressed). The \texttt{scroll-pre} and \texttt{scroll-post} boxes sit 1--2 orders of magnitude above the \texttt{SEARCH} boxes in every combination.}
\label{fig:country_age_phase_boxplot}
\end{figure*}

\paragraph{Phase-level engagement gap}
Across all three countries, For-You-feed items surface substantially higher play counts than search-surfaced items, typically by an order of magnitude on the median (Figure~\ref{fig:country_age_phase_boxplot}). We attribute this to the different economic logics of the two sub-systems: the For-You feed optimizes for watch-time virality, while search retrieves a long-tail query-matched pool. The effect is stable across ages and shows up at both \texttt{scroll-pre} and \texttt{scroll-post}, suggesting it is not an artifact of the probe. The full passive phase, collected on the same accounts at an earlier window, mirrors the active-phase scrolling pattern: mean play counts are again in the millions per video (Table~\ref{tab:passive_engagement}).

\begin{table}[h]
\centering
\small
\begin{tabular}{llrrr}
\toprule
Country & Age & Videos & Mean plays & Mean likes \\
\midrule
FR & 13 & 2{,}498 & 5.14 M & 443 K \\
FR & 16 & 2{,}071 & 6.50 M & 613 K \\
FR & 19 & 2{,}760 & 4.60 M & 389 K \\
FR & 40 & 2{,}159 & 6.71 M & 569 K \\
IT & 13 & 1{,}841 & 3.93 M & 256 K \\
IT & 16 & 1{,}805 & 5.73 M & 496 K \\
IT & 19 & 1{,}975 & 1.88 M & 183 K \\
IT & 40 & 1{,}709 & 5.50 M & 456 K \\
SW & 13 & 2{,}195 & 6.24 M & 737 K \\
SW & 16 & 2{,}019 & 6.65 M & 709 K \\
SW & 19 & 2{,}019 & 4.78 M & 456 K \\
SW & 40 & 1{,}981 & 5.38 M & 712 K \\
\bottomrule
\end{tabular}
\caption{Passive-phase engagement statistics per (country, age) combination, over all item occurrences. Companion to Table~\ref{tab:dataset_detail} (active phase).}
\label{tab:passive_engagement}
\end{table}

\begin{table*}[t]
\centering
\small
\begin{tabular}{llrrrrr}
\toprule
Phase & Country & Native & English & Other & Unknown & Total \\
\midrule
\multirow{3}{*}{Passive}
 & FR & 11.6\% & 45.7\% & 27.3\% & 15.4\% & 6{,}842 \\
 & IT & 20.6\% & 30.3\% & 26.7\% & 22.3\% & 5{,}344 \\
 & SW & 10.5\% & 48.3\% & 26.1\% & 15.1\% & 5{,}922 \\
\midrule
\multirow{3}{*}{Active}
 & FR & 37.9\% & 25.5\% & 26.8\% &  9.8\% & 7{,}130 \\
 & IT & 38.7\% & 23.9\% & 24.5\% & 12.8\% & 7{,}531 \\
 & SW & 30.1\% & 28.9\% & 31.1\% &  9.9\% & 8{,}217 \\
\bottomrule
\end{tabular}
\caption{Detected-language distribution of unique videos per phase and persona country. ``Native'' is French for FR, Italian for IT, Swedish for SW; ``Other'' aggregates remaining \texttt{langdetect} labels; ``Unknown'' marks captions under 8 characters or unscoreable.}
\label{tab:lang_overview}
\end{table*}

\section{Stage-2 Supplementary Figures and Tables}
\label{app:stage2_supp}

This appendix collects the Stage-2 figures and tables that are referenced from \S\ref{sec:results:prevalence}--\S\ref{sec:results:failures} but moved out of the main text for space.

\paragraph{Provider-block worst-case bound}
Counting every block-or-parse-fail item (96 of $5{,}323$ E3 inputs, $1.80\%$) as either \textsc{Harmful} or \textsc{Not Harmful} yields per-country headline intervals of $[26.5\%, 28.4\%]$ for France (vs.\ reported $27.0\%$), $[35.0\%, 36.5\%]$ for Italy ($35.5\%$), and $[27.5\%, 29.5\%]$ for Sweden ($28.0\%$); the IT-16 case shifts at most from $40.4\%$ to $[39.9\%, 41.1\%]$. The IT~$>$~SW~$\geq$~FR ordering and the IT-13 / IT-16 ceiling-effect finding both survive this bound.

\begin{table}[h]
\centering
\small
\begin{tabular}{lrrr}
\toprule
Country & E2 blocks & E3 blocks & Block rate \\
\midrule
FR  & 31 & 26 & $\approx 1.5\%$ \\
IT  & 11 & 11 & $\approx 0.7\%$ \\
SW  & 15 & 27 & $\approx 1.2\%$ \\
\midrule
All & 57 & 64 & $\approx 1.1\%$ \\
\bottomrule
\end{tabular}
\caption{Stage-2 Gemini \texttt{PROHIBITED\_CONTENT} counts per country and modality, over ${\sim}5{,}300$ inputs per modality. Per-category breakdown in \S\ref{sec:results:failures}.}
\label{tab:stage2_blocks_summary}
\end{table}

\paragraph{Sample top-up details}
The initial $10\%$ phase-stratified draw was $3{,}862$ items; after excluding $294$ unavailable MP4s, $3{,}568$ were fed to Gemini under both E2 and E3. The \texttt{scroll-pre} and \texttt{scroll-post} cases held ${\sim}17$ records each, versus ${\sim}150$ for \texttt{SEARCH}, which was too thin to anchor the within-account \texttt{SEARCH}/\texttt{scroll-pre} comparison in \S\ref{sec:results:phase}. We therefore topped up both phases from the same accounts, days, and FYP snapshots until each \texttt{scroll-pre} and \texttt{scroll-post} case held ${\sim}100$ items, taking the final sample to $5{,}248$ usable E2 verdicts and $5{,}227$ E3 verdicts ($5{,}108$ paired).

\paragraph{Original-draw ratio robustness}
We recompute the \S\ref{sec:results:phase} \texttt{SEARCH}/\texttt{scroll-pre} ratios on the \emph{original} 10\% phase-stratified draw before the top-up. Across the ten combinations where \texttt{scroll-pre} is positive in both draws, the original-draw factor is $1.2\times$ to $7.4\times$ (vs.\ $1.0\times$ to $8.1\times$ topped-up); IT-19 and IT-40 are undefined on the original draw because the small $n_{\text{scroll-pre}} \in [15, 17]$ produced zero harmful items, which is the noise regime the top-up was designed to escape.

\paragraph{Account-clustered bootstrap intervals}
Table~\ref{tab:clustered_ci} reports the \S\ref{sec:results:phase} active-phase harm rates with $95\%$ CIs from a hierarchical bootstrap that resamples the three accounts per cell with replacement and then resamples videos within each drawn account, so that between-account correlation is preserved.
Account attribution covers $99.3$--$99.8\%$ of Stage-2 active records per phase; a video captured by two accounts contributes to both accounts' pools.
The clustered intervals match the video-level ones closely (median width ratio $1.00$ across the 36 E3 cells, maximum $1.85\times$), reflecting the low between-account variability visible in the per-account \texttt{SEARCH} rate ranges.

\begin{table}[h]
\centering
\small
\setlength{\tabcolsep}{3pt}
\begin{tabular}{lrrc}
\toprule
Cell & \texttt{scroll-pre} \% & \texttt{SEARCH} \% & acct.\ range \\
\midrule
FR-13 & 9.5 [2.9, 19.2] & 41.5 [36.8, 50.0]$^{*}$ & [41.4, 45.3] \\
FR-16 & 18.9 [13.7, 29.0] & 49.2 [47.1, 61.7]$^{*}$ & [51.5, 56.9] \\
FR-19 & 6.7 [1.7, 10.7] & 50.4 [43.9, 59.9]$^{*}$ & [49.0, 55.9] \\
FR-40 & 21.8 [14.8, 32.7] & 55.6 [48.3, 65.8]$^{*}$ & [50.0, 62.1] \\
IT-13 & 44.2 [34.0, 58.4] & 43.1 [38.2, 51.7] & [42.7, 46.7] \\
IT-16 & 34.4 [23.9, 41.9] & 45.7 [38.6, 53.6] & [44.4, 49.3] \\
IT-19 & 6.9 [1.1, 12.5] & 50.4 [44.8, 58.0]$^{*}$ & [49.4, 54.0] \\
IT-40 & 7.5 [1.0, 13.6] & 49.2 [42.9, 57.8]$^{*}$ & [47.1, 53.5] \\
SW-13 & 12.5 [4.7, 18.3] & 36.7 [30.9, 44.9]$^{*}$ & [33.3, 40.0] \\
SW-16 & 16.8 [11.2, 29.3] & 37.3 [34.2, 46.8]$^{*}$ & [39.0, 41.5] \\
SW-19 & 23.2 [10.5, 27.2] & 34.6 [30.2, 43.1]$^{*}$ & [34.8, 39.3] \\
SW-40 & 11.1 [3.2, 16.7] & 43.7 [35.2, 50.7]$^{*}$ & [38.8, 46.6] \\
\bottomrule
\end{tabular}
\caption{Gemini~E3 harm rates with $95\%$ account-clustered hierarchical-bootstrap CIs, per active-phase cell. $^{*}$ marks \texttt{SEARCH} intervals disjoint from the cell's \texttt{scroll-pre} interval. The last column is the range of per-account \texttt{SEARCH} harm rates within the cell.}
\label{tab:clustered_ci}
\end{table}

\paragraph{Policy-tier split}
Table~\ref{tab:age_policy_split} decomposes each cell's harm rate into the 18+-restricted tier (Sexually Suggestive, Nudity and Body Exposure, Alcohol/Tobacco/Drugs, Gambling, Shocking and Graphic) and the universally prohibited tier (the remaining eight categories), as discussed in \S\ref{sec:results:phase}.
The Disordered Eating placement is arguable (promotion is prohibited, generic weight-management content is 18+-restricted); moving it between tiers shifts no cell by more than $2$~pp.

\begin{table}[h]
\centering
\small
\begin{tabular}{lrrrr}
\toprule
 & \multicolumn{2}{c}{Passive} & \multicolumn{2}{c}{\texttt{SEARCH}} \\
\cmidrule(lr){2-3}\cmidrule(lr){4-5}
Cell & 18+ & Proh. & 18+ & Proh. \\
\midrule
FR-13 & 12.2 & 12.2 & 26.7 & 14.8 \\
FR-16 & 16.9 & 6.5 & 28.7 & 20.5 \\
FR-19 & 17.0 & 5.5 & 39.1 & 11.3 \\
FR-40 & 13.4 & 9.2 & 36.3 & 19.3 \\
IT-13 & 17.2 & 10.7 & 34.6 & 8.5 \\
IT-16 & 33.3 & 9.7 & 32.8 & 12.9 \\
IT-19 & 45.9 & 2.7 & 34.8 & 15.6 \\
IT-40 & 26.7 & 10.5 & 36.3 & 12.9 \\
SW-13 & 8.6 & 15.1 & 24.4 & 12.2 \\
SW-16 & 21.1 & 14.6 & 25.9 & 11.4 \\
SW-19 & 27.8 & 10.5 & 19.9 & 14.7 \\
SW-40 & 24.5 & 5.5 & 25.9 & 17.8 \\
\bottomrule
\end{tabular}
\caption{Percentage-point contribution of the 18+-restricted and universally prohibited category tiers to each cell's Gemini~E3 harm rate, passive phase and \texttt{SEARCH} phase.}
\label{tab:age_policy_split}
\end{table}

\paragraph{Decomposing the Italian lead}
Three passive-phase decompositions support the reading in \S\ref{sec:discussion}.
By category, Sexually Suggestive content contributes $23.8$~pp of Italy's $39.0\%$ passive rate versus $12.2$~pp in France and $15.8$~pp in Sweden, so a single category accounts for most of the cross-country gap.
By content language (Table~\ref{tab:passive_lang_harm}), Italian-language videos are flagged at $46.8\%$, but English-language videos served to Italian accounts are also flagged at $32.4\%$, clearly above the English-language rate in France ($19.5\%$), with Sweden in between ($27.5\%$).
By cross-country circulation, passive videos that also appear in another country's corpus are flagged at $19.2\%$ when served to Italian accounts, statistically indistinguishable from the shared-content rate in France ($26.4\%$) and Sweden ($28.3\%$), while Italy-exclusive content is flagged at $41.7\%$; the Italian lead is carried entirely by content that circulates only in the Italian pool.

\begin{table}[h]
\centering
\small
\setlength{\tabcolsep}{4pt}
\begin{tabular}{lrrrr}
\toprule
Country & Native & English & Other & Unknown \\
\midrule
FR & 21.9 (73) & 19.5 (261) & 27.5 (149) & 28.1 (89) \\
IT & 46.8 (94) & 32.4 (136) & 35.1 (114) & 46.1 (89) \\
SW & 51.1 (45) & 27.5 (258) & 27.1 (133) & 44.9 (69) \\
\bottomrule
\end{tabular}
\caption{Passive-phase Gemini~E3 harm rate (\%) by detected content language, with $n$ in parentheses. ``Native'' is the persona country's language; detection follows the caption-based protocol of Table~\ref{tab:lang_overview}.}
\label{tab:passive_lang_harm}
\end{table}

\begin{figure}[h]
\centering
\includesvg[width=\linewidth]{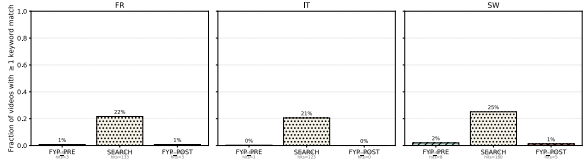}
\caption{Fraction of Stage-2 active videos whose caption $\cup$ transcript matches at least one of the country's $21$ search keywords, by phase.}
\label{fig:stage2_keyword_match}
\end{figure}

\paragraph{Keyword-coverage baseline}
Figure~\ref{fig:stage2_keyword_match} reports the fraction of Stage-2 active videos whose caption and transcript together contain at least one of the country's 21 harm keywords (whole-word, case- and diacritic-insensitive match).
\texttt{SEARCH} videos contain a literal keyword in $20$--$25\%$ of cases across all three countries (FR $21.6\%$, IT $20.5\%$, SW $25.1\%$), while the \texttt{scroll-pre} and \texttt{scroll-post} baselines sit at $0$--$2\%$.
The two-order-of-magnitude gap is a textual sanity check on the probe; the remaining $75$--$80\%$ of \texttt{SEARCH} videos are surfaced by the platform's own topic-and-engagement match around the query rather than by literal keyword presence in caption or transcript.

\begin{table}[h]
\centering
\small
\begin{tabular}{lrrrr}
\toprule
Country & $n$ paired & E2 rate & E3 rate & $\kappa$ \\
\midrule
FR  & 1{,}768 & 0.309 & 0.270 & 0.604 \\
IT  & 1{,}602 & 0.385 & 0.346 & 0.633 \\
SW  & 1{,}738 & 0.318 & 0.278 & 0.599 \\
\midrule
All & 5{,}108 & 0.336 & 0.297 & 0.614 \\
\bottomrule
\end{tabular}
\caption{Stage-2 cross-modality agreement between E2 and E3 on the binary harm verdict ($n = 5{,}108$ paired).}
\label{tab:stage2_e2e3}
\end{table}

\begin{figure*}[h]
\centering
\includesvg[width=0.95\textwidth]{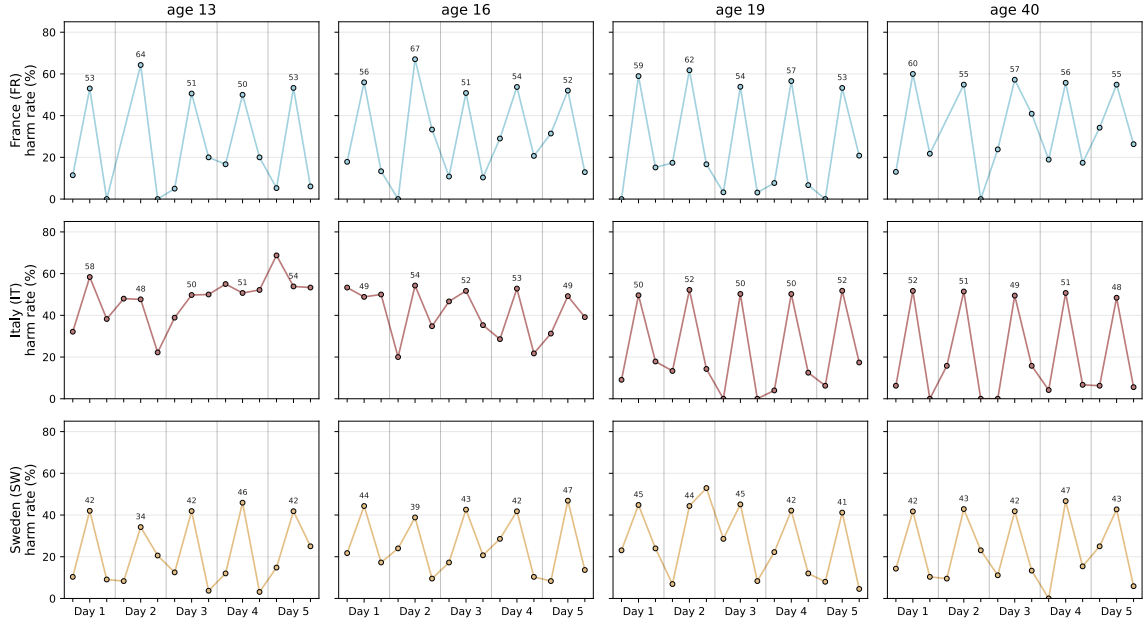}
\caption{Stage-2 Gemini~E3 daily harm rate per (country, age) over the five active-collection days. Rows are country, columns are age; each major x-tick is one collection day, with three points per day in order \texttt{scroll-pre} (scroll) $\to$ \texttt{SEARCH} $\to$ \texttt{scroll-post} (scroll). The triangular per-day shape is the local within-day \texttt{SEARCH} lift.}
\label{fig:stage2_daily_active}
\end{figure*}

\begin{table*}[h]
\centering
\small
\begin{tabular}{lrrrr}
\toprule
Harm subcategory & $n$ & both~H & E2~only & E3~only \\
\midrule
Sexually Suggestive Content        & 623 & 395 & 133 &  95 \\
Shocking and Graphic Content       & 145 &  79 &  47 &  19 \\
Disordered Eating, Body Image      & 173 & 113 &  34 &  26 \\
Dangerous Activity, Challenges     &  95 &  40 &  28 &  27 \\
Harassment and Bullying            & 104 &  55 &  20 &  29 \\
Alcohol, Tobacco, and Drugs        & 143 &  96 &  27 &  20 \\
Gambling                           & 101 &  74 &  13 &  14 \\
Nudity and Body Exposure           &  60 &  41 &  15 &   4 \\
Integrity and Authenticity         &  39 &  19 &  13 &   7 \\
Sexual and Physical Abuse          &  49 &  33 &  13 &   3 \\
Hate Speech or Behavior            &  34 &  22 &   6 &   6 \\
Suicide and Self-Harm              &  25 &  15 &   6 &   4 \\
\midrule
Total flagged by at least one      & 1{,}591 & 982 & 355 & 254 \\
\bottomrule
\end{tabular}
\caption{Per-subcategory disagreement between E2 (native video) and E3 (primary) on the $3{,}417$ paired Stage-2 items. \emph{both~H} = both flagged harmful with this subcategory; \emph{E2~only} = E2 flagged, E3 did not; \emph{E3~only} = E3 flagged, E2 did not.}
\label{tab:stage2_e2e3_per_cat}
\end{table*}

\begin{figure}[h]
\centering
\includesvg[width=\linewidth]{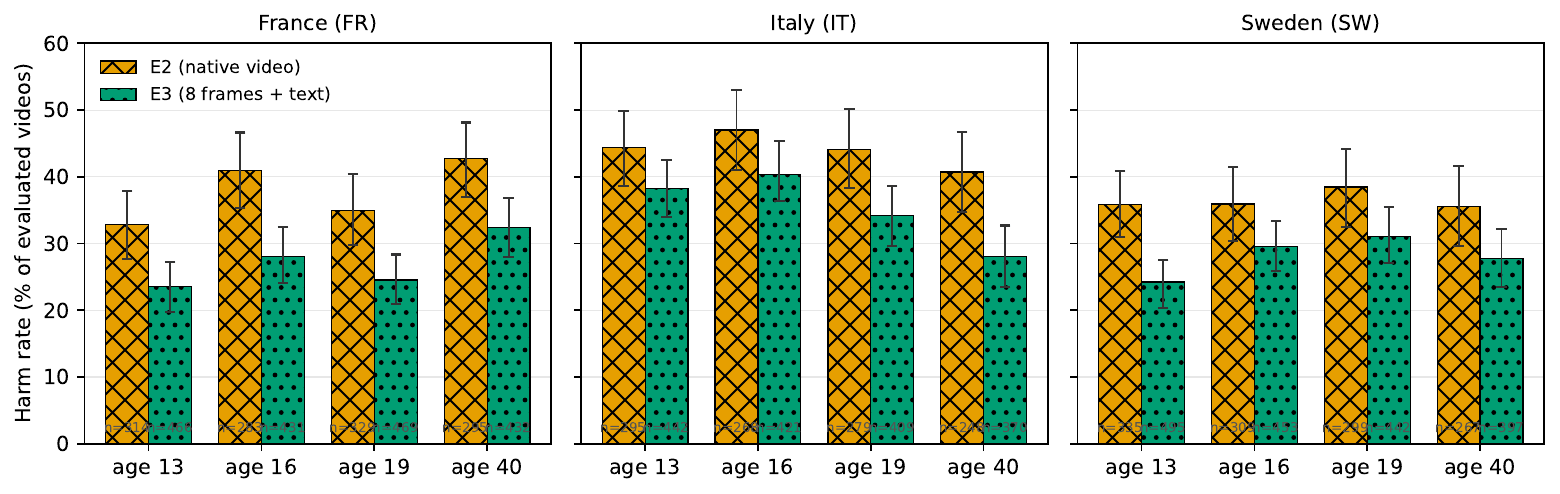}
\caption{Stage-2 harm rate per (country, age) combination on Gemini~E3 and native-video E2, with $95\%$ video-level bootstrap CIs. Italy carries the highest E3 rate on every age persona; the within-country age gradient differs across the three countries. The cleaner passive-only view is in the main text (Figure~\ref{fig:stage2_harm_passive_e3}, \S\ref{sec:results:passive}).}
\label{fig:stage2_harm_country_age}
\end{figure}

\begin{figure}[h]
\centering
\includesvg[width=\linewidth]{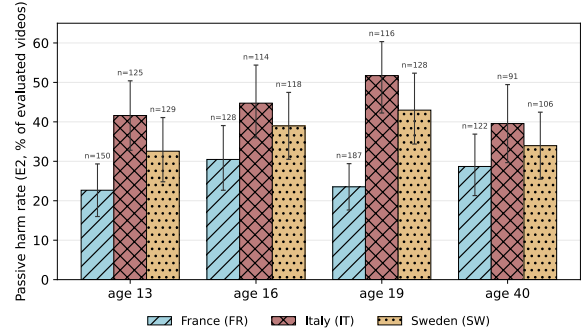}
\caption{Stage-2 harm rate per (country, age) on the passive collection phase (FYP scrolling only, no search probe) under Gemini~E2 (native video); companion to Figure~\ref{fig:stage2_harm_passive_e3}.}
\label{fig:stage2_harm_passive_e2}
\end{figure}

\section{Stage-1 Diagnostic Detail}
\label{app:stage1_detail}

This appendix collects the per-condition diagnostics referenced from \S\ref{sec:results:agreement}--\S\ref{sec:results:failures}. The numbers are computed on the 300-video two-annotator-with-resolution subset (299 binary reference verdicts, $80$--$99$ per combination depending on model failures) and are pre-Stage-2. The XLM-R supervised text baseline is included alongside the four MLLM families.

\paragraph{Inter-annotator agreement and resolution}
Two native-speaker annotators per country independently labeled the 300-video Stage-1 subset.
Before resolution, the two annotators agreed on the binary harm verdict on $80.3\%$ of items ($n = 300$); Cohen's $\kappa$ on the binary verdict is $0.54$ in the aggregate, with per-country $\kappa_{\text{FR}} = 0.42$, $\kappa_{\text{IT}} = 0.65$, $\kappa_{\text{SW}} = 0.48$.
All $59$ binary-verdict disagreements were settled in a joint resolution session that produced a single consensus label per video.
A further resolution rule was applied at analysis time without re-soliciting the annotators: when both annotators rated a video \textsc{Harmful} but disagreed on its primary subcategory, the consensus row carries the union of the two subcategory choices.
The resulting per-video table is the \textit{final reference} against which the LLM agreement numbers in this appendix are computed.

\begin{figure}[t]
\centering
\includesvg[width=0.95\linewidth]{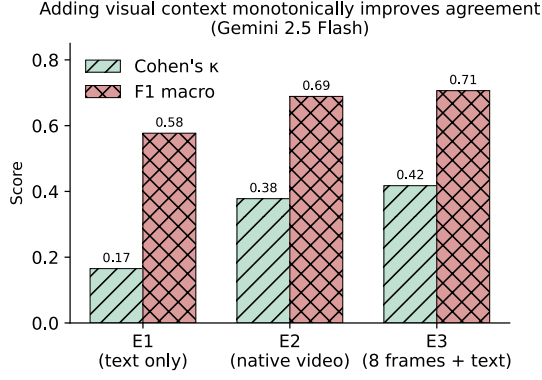}
\caption{Cohen's $\kappa$ and macro-$F_1$ across the three input conditions for Gemini 2.5 Flash on the 300-video Stage-1 subset. E3 (eight frames) is the strongest at ${\sim}2\times$ lower per-call cost than E2 (native video).}
\label{fig:stage1_modality_monotonicity}
\end{figure}

\paragraph{Modality monotonicity}
Figure~\ref{fig:stage1_modality_monotonicity} shows how Cohen's $\kappa$ and macro-$F_1$ move across the three input conditions for the winning configuration (Gemini 2.5 Flash) on the 300-video Stage-1 subset; the per-condition numerics are cited inline in \S\ref{sec:results:agreement}.
On the same panel, Qwen3-VL-32B does \emph{not} show the same monotonic pattern, sitting flat between E2 and E3 at $\kappa \approx 0.18$, which suggests its native-video pathway and its frame-based pathway converge on the same (weak) judgment rather than complementing one another.

\begin{figure}[t]
\centering
\includesvg[width=0.85\linewidth]{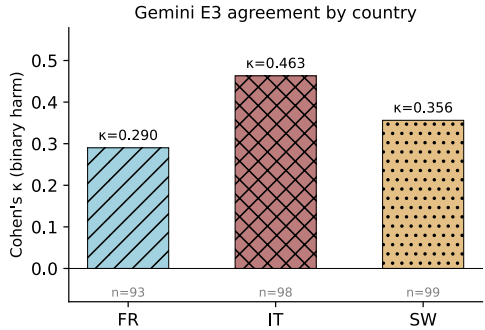}
\caption{Per-country Cohen's $\kappa$ for the winning configuration (Gemini~2.5~Flash, E3).}
\label{fig:stage1_country_kappa}
\end{figure}

\paragraph{Per-country agreement of the winner}
Figure~\ref{fig:stage1_country_kappa} expands the per-country $\kappa$ ranking of the Stage-1 winning configuration referenced in \S\ref{sec:results:agreement}. Per-country values are $\kappa_{\text{FR}} = 0.290$ (CI $[0.03, 0.52]$, $n_{\text{FR}} = 93$), $\kappa_{\text{IT}} = 0.463$ (CI $[0.29, 0.63]$, $n_{\text{IT}} = 98$), and $\kappa_{\text{SW}} = 0.356$ (CI $[0.16, 0.54]$, $n_{\text{SW}} = 99$): Italian content reaches the highest agreement and French the lowest, which inverts the naive prediction that higher per-language moderator allocation (per \citet{tonneau2025languagemoderation}'s data) should correlate with higher MLLM-vs-annotator agreement at the point-estimate level; the per-country CIs are wide and pairwise overlapping, so the inversion is consistent with sampling variation at this $n$ rather than firm evidence against the hypothesis. France contributes the largest share of provider-blocked items (Table~\ref{tab:stage1_failures}), so part of the FR$\kappa$ gap is attributable to the removal of those items from the comparison.

A complementary decomposition uses the pre-resolution two-annotator $\kappa$ per country as a reference-quality ceiling: the LLM cannot agree with the consensus better than the consensus agrees with itself. The pre-resolution two-annotator $\kappa$ values are $\kappa^{\text{ann}}_{\text{FR}} = 0.42$, $\kappa^{\text{ann}}_{\text{IT}} = 0.65$, $\kappa^{\text{ann}}_{\text{SW}} = 0.48$. Gemini~E3 reaches $69\%$ of this ceiling on French content ($0.290 / 0.42$), $71\%$ on Italian ($0.463 / 0.65$), and $74\%$ on Swedish ($0.356 / 0.48$). The fact that the ratio is roughly constant across the three countries indicates that the absolute per-country $\kappa$ gap (FR $< $ SW $< $ IT) tracks the per-country reference-quality gap rather than a country-specific model deficit, and that the cross-country $\kappa$ inversion of the moderator-allocation prediction is best read as a reference-quality artifact: French annotator disagreement is the main driver of $\kappa_{\text{FR}}$ being the lowest model-vs-reference combination.

\paragraph{Per-country precision/recall recalibration of Stage-2 rates}
Stage-1 per-country precision and recall on the binary harm verdict (FR: $P/R = 0.50 / 0.33$, $P/R \approx 1.50$; IT: $0.80 / 0.60$, $\approx 1.34$; SW: $0.62 / 0.51$, $\approx 1.21$) give a country-specific true-positive-rate correction for the Stage-2 harm rates of \S\ref{sec:results:prevalence}. Applied uniformly across (country, age) combinations, the recalibration leaves the cross-country ordering intact at ages 13, 16, and 19 (Italy remains highest) but inverts the age-$40$ case, where corrected FR-$40$ ($\approx 49\%$) overtakes IT-$40$ ($\approx 38\%$) and SW-$40$ ($\approx 33\%$). The Stage-1 $P/R$ values are themselves point estimates on per-country $n \in [93, 99]$, so the inversion is consistent with both ``IT-$40$ ties FR-$40$'' and ``IT highest on every age''; we report it as a caveat on the strongest reading of the cross-country claim at age 40 rather than a re-ordering.

\begin{figure}[t]
\centering
\includesvg[width=\linewidth]{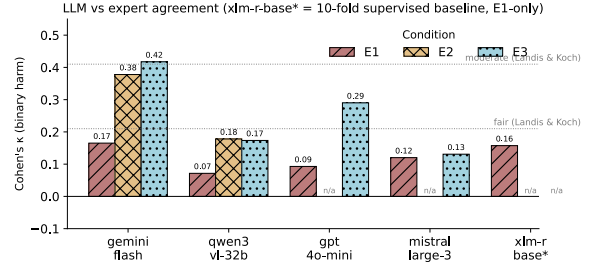}
\caption{Aggregate Cohen's $\kappa$ on the binary harm verdict across the four MLLMs and the XLM-R supervised baseline, by input condition. GPT-4o-mini and Mistral are not natively video-capable and are omitted from E2. Dotted lines: $\kappa = 0.21$ (fair) and $\kappa = 0.41$ (moderate) \citep{landis1977kappa}.}
\label{fig:stage1_kappa_grid}
\end{figure}

\paragraph{Aggregate $\kappa$ grid}
Figure~\ref{fig:stage1_kappa_grid} collapses the three-panel split of Figure~\ref{fig:stage1_kappa_grid_by_country} into a single aggregate model$\times$condition grid for completeness. The aggregate ordering averages over the strong cross-lingual asymmetry visible in the per-country split: the French panel is the only one where Qwen~E2 exceeds Qwen~E3 ($\kappa_{\text{FR}} = 0.31$ vs.\ $0.08$); the Italian panel concentrates the moderate-agreement combinations in the roster (Gemini~E2 reaches $\kappa_{\text{IT}} = 0.51$, Gemini~E3 reaches $\kappa_{\text{IT}} = 0.46$, and GPT-4o-mini~E3 separately reaches $\kappa_{\text{IT}} = 0.41$); and the Swedish panel collapses to Gemini~E3 as the strongest non-Italian combination (non-Gemini combinations on Swedish content stay at or below $\kappa = 0.17$ across all three input conditions, and Qwen~E1 and GPT~E1 sit at or below chance).

\begin{figure*}[t]
\centering
\includesvg[width=\textwidth]{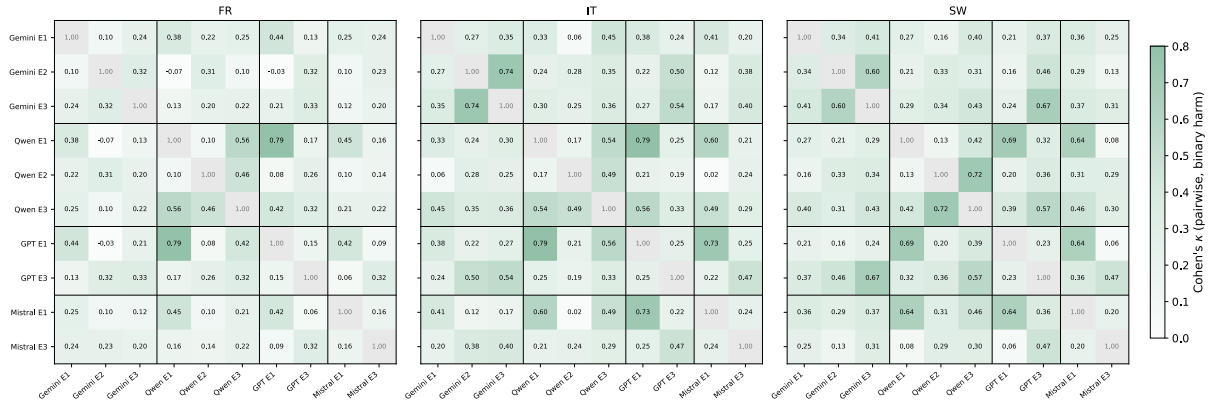}
\caption{Pairwise Cohen's $\kappa$ between every pair of the eleven Stage-1 (model, condition) combinations (the ten MLLM combinations plus the XLM-R baseline), split by persona country. Off-diagonals are computed on each pair's intersection of paired \texttt{video\_id}s within the country slice (per-pair $n \approx 95$--$108$).}
\label{fig:stage1_intermodel}
\end{figure*}

\paragraph{Pairwise inter-model agreement, per country}
Figure~\ref{fig:stage1_intermodel} reports Cohen's $\kappa$ between every pair of the eleven Stage-1 (model, condition) combinations (the ten MLLM combinations plus the XLM-R baseline) on the binary harm label, split by persona country and computed on each pair's intersection of paired \texttt{video\_id}s within the country slice (per-combination $n \approx 95$--$108$).
The same three blocks dominate every country panel: (i) a text-only cluster among Qwen~E1, GPT~E1, and Mistral~E1, with the Qwen$\leftrightarrow$GPT pair the strongest in every country; (ii) a frames-and-video cluster anchored by Gemini~E2$\leftrightarrow$Gemini~E3 and Gemini~E3$\leftrightarrow$GPT~E3; and (iii) an isolated Qwen~E2 row that agrees only weakly with everything outside its own provider, consistent with the high \texttt{413~RequestTooLarge} block rate documented in Table~\ref{tab:stage1_failures}.
Within-model, comparing Qwen~E1 to Qwen~E3 and GPT~E1 to GPT~E3 makes the effect of adding frames visible: GPT-4o-mini's verdicts change substantially when shown frames, while Qwen3-VL-32B's do not.
Read together with Figure~\ref{fig:stage1_kappa_grid_by_country}, the per-country matrices support the language-prior account: in conditions where the verdict is text-driven, the four model families agree with each other at moderate-to-substantial levels regardless of provider, so the disagreement with the annotators comes mostly from the visual-judgment side of the task, not from the text-classification side.

\paragraph{Confusion of the winner}
The consensus-reference-vs-Gemini-E3 confusion over the $n = 290$ paired records shows that the model is conservative on harm: it under-flags more than it over-flags (48 false negatives vs.\ 24 false positives), inverting the usual concern about LLM over-moderation \citep{kumar2024watchyourlanguage} and consistent with Gemini's safety-layer bias toward refusing rather than mislabeling sensitive content. On the 52 jointly-harmful items where both reference and model say \textsc{Harmful}, agreement on the \emph{primary} subcategory is 29\% (15/52); under a relaxed definition that scores the model correct when its primary prediction matches either the human primary or human secondary subcategory, agreement rises to 54\% (28/52). The bulk of strict-match disagreements concentrate on adjacent categories within the same harm domain (nudity-vs-suggestive, dangerous-challenge-vs-shocking) rather than across harm domains.

\begin{figure}[t]
\centering
\includesvg[width=\linewidth]{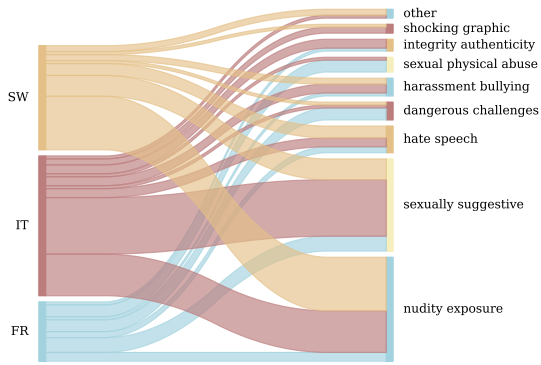}
\caption{Country~$\to$~harm-subcategory mix for the 102 reference-flagged harmful videos. Ribbon widths are proportional to absolute counts.}
\label{fig:stage1_category_sankey}
\end{figure}

\paragraph{Country-specific harm-category mix}
Figure~\ref{fig:stage1_category_sankey} renders the country-conditional distribution of harm subcategories among the reference-flagged harmful videos (102 across FR/IT/SW). The Sankey routes the within-country normalized category shares as flows, which keeps the absolute counts per country visible alongside the cross-country comparison. France's harmful set is dominated by \textit{dangerous activity and challenges}; Italy and Sweden are dominated by \textit{nudity and body exposure} together with \textit{sexually suggestive} content (roughly two-thirds of harmful items in each), with hate-speech and harassment categories more visible in Sweden than elsewhere. A single global $\kappa$ figure averages over these different harm distributions, so Stage-2 must report per-country, per-category prevalence side by side with overall agreement to avoid hiding the asymmetry.

\paragraph{Winner vs.\ runner-up table}
Table~\ref{tab:stage1_winner} reports the full Stage-1 metrics for Gemini~E3 and the strongest open-weight comparator under the same condition. The 0.27-point $\kappa$ gap between the two configurations is large enough that no plausible scaling discount overturns the choice of Gemini~E3 as the Stage-2 configuration.

\begin{table}[t]
\centering
\small
\begin{tabular}{lrr}
\toprule
Metric & Gemini~E3 & Qwen3-VL-32B E3 \\
\midrule
$n$ paired              & 290   & 292   \\
Cohen's $\kappa$        & 0.417 & 0.173 \\
Accuracy                & 0.752 & 0.682 \\
$F_1$ harmful           & 0.591 & 0.340 \\
$F_1$ not-harmful       & 0.822 & 0.790 \\
Macro-$F_1$             & 0.706 & 0.565 \\
Precision (harmful)     & 0.684 & 0.571 \\
Recall (harmful)        & 0.520 & 0.242 \\
\bottomrule
\end{tabular}
\caption{Stage-1 metrics for the winning configuration and the strongest open-weight comparator under the same E3 condition.}
\label{tab:stage1_winner}
\end{table}

\paragraph{Precision and recall across all Stage-1 combinations}
Table~\ref{tab:stage1_full_metrics} reports the full precision/recall/F$_1$ breakdown on the binary harm verdict for every Stage-1 (model, condition) combination against the two-annotator final reference. Gemini~E3 leads the table on every harm-side metric; the supervised XLM-R baseline beats all zero-shot LLMs on harmful-class recall but trails on precision, consistent with a supervised model fitting the positive class on a small training set. All four non-Gemini frame and text conditions sit at or below $50\%$ recall on harmful, which is the largest single failure mode of the LLM auditors.

\begin{table}[t]
\centering
\small
\setlength{\tabcolsep}{3.5pt}
\begin{tabular}{lrrrrr}
\toprule
Configuration & P$_{\textsc{H}}$ & R$_{\textsc{H}}$ & F$_1^{\textsc{H}}$ & F$_1^{\textsc{NH}}$ & mF$_1$ \\
\midrule
Gemini~E3 (frames)        & \textbf{0.68} & 0.52 & \textbf{0.59} & \textbf{0.82} & \textbf{0.71} \\
Gemini~E2 (native)        & 0.59 & \textbf{0.58} & 0.58 & 0.80 & 0.69 \\
GPT-4o-mini E3 (frames)   & 0.56 & 0.47 & 0.51 & 0.78 & 0.64 \\
Qwen E2 (native)          & 0.60 & 0.24 & 0.35 & 0.78 & 0.56 \\
Qwen E3 (frames)          & 0.57 & 0.24 & 0.34 & 0.79 & 0.57 \\
Gemini~E1 (text)          & 0.49 & 0.34 & 0.40 & 0.76 & 0.58 \\
XLM-R E1 (supervised)     & 0.43 & 0.49 & 0.46 & 0.70 & 0.58 \\
Mistral E3 (frames)       & 0.43 & 0.44 & 0.43 & 0.70 & 0.57 \\
Mistral E1 (text)         & 0.51 & 0.20 & 0.29 & 0.78 & 0.53 \\
GPT-4o-mini E1 (text)     & 0.54 & 0.13 & 0.21 & 0.79 & 0.50 \\
Qwen E1 (text)            & 0.50 & 0.12 & 0.19 & 0.79 & 0.49 \\
\bottomrule
\end{tabular}
\caption{Stage-1 precision (P$_{\textsc{H}}$), recall (R$_{\textsc{H}}$), per-class $F_1$, and macro-$F_1$ on the binary harm verdict, across the eleven configurations (ten MLLMs plus the XLM-R baseline) against the two-annotator final reference. Rows sorted by Cohen's $\kappa$; per-column maxima in \textbf{bold}.}
\label{tab:stage1_full_metrics}
\end{table}

\paragraph{Supervised XLM-R baseline}
The XLM-R row in Figure~\ref{fig:stage1_kappa_grid} and Table~\ref{tab:stage1_full_metrics} is a 10-fold repeated stratified train/test cross-validation of \texttt{xlm-roberta-base} on the same 300-video Stage-1 subset used to evaluate the four MLLMs. The input is the E1 text only, caption plus audio transcript, exactly the field shown to the text-only LLM runs, and the target is the consensus binary harm verdict. Each fold holds out $19\%$ of items as test, carves $15\%$ of the remaining train side as a dev split for early stopping, and trains for up to $10$ epochs with AdamW (lr $2\!\times\!10^{-5}$, batch size $16$, max sequence length $256$, weight decay $0.01$, linear warmup over $6\%$ of steps, early-stopping patience of $4$ epochs on dev macro-$F_1$). Class imbalance is handled by inverse-frequency reweighting computed on the train side. Folds use random seeds $0$--$9$ and the per-video predicted probability is the mean across the (typically two) folds in which a given video appears in the test split; the metrics in Table~\ref{tab:stage1_full_metrics} are computed on those aggregated per-video predictions against the same final reference. We read the baseline as a sanity-check anchor for the LLM E1 combinations rather than as a competitive system: at this training-set size ($n \approx 242$ per fold after dropping \textsc{Not Available} items), the supervised model can beat zero-shot LLMs on harmful-class recall but cannot match Gemini~E3's precision-recall balance.

\paragraph{Provider-block counts}
\begin{table}[!ht]
\centering
\small
\begin{tabular}{lrrr}
\toprule
Failure family & E1 & E2 & E3 \\
\midrule
Gemini \texttt{PROHIBITED\_CONTENT}    & 3 & 6 & 2 \\
Gemini response-parse failure           & 0 & 1 & 5 \\
Qwen \texttt{413 RequestTooLarge}      & 0 & 14 & 0 \\
Qwen \texttt{data\_inspection\_failed} & 0 & 2 & 3 \\
GPT-4o-mini OpenRouter credit-cap      & 8 & n/a & 15 \\
\bottomrule
\end{tabular}
\caption{Provider-side refusals on the 300-video Stage-1 subset, by experiment. Each cell counts videos where the model's response could not be evaluated because the provider blocked the call or returned an unparseable response.}
\label{tab:stage1_failures}
\end{table}

The Qwen \texttt{413 RequestTooLarge} dominates absolute volume because DashScope's native-video endpoint enforces a per-call payload cap; the workaround would change what Qwen sees and break comparability with unmodified Gemini~E2, so we report Qwen~E2 only as a comparator. Gemini's blocks are policy decisions; they could be relaxed via \texttt{safety\_settings} = \textsc{BLOCK\_NONE}, appropriate only under formal ethics approval.

\end{document}